\documentclass[sn-apa,iicol]{sn-jnl}% APA Reference Style
\usepackage{verbatim,color,amssymb}
\usepackage{amsmath}					
\usepackage{amsthm}					
\usepackage{natbib}
\usepackage{multirow}
\usepackage{setspace}
\usepackage[mathscr]{euscript}
\usepackage{fancyhdr}
\usepackage{enumitem}
\usepackage{graphicx}
\usepackage{geometry}
\usepackage{bm}
\usepackage{dsfont}
\usepackage{tabularx, booktabs}
\usepackage{lscape}
\usepackage{subfig}
\usepackage{caption}
\usepackage{float}
\usepackage{setspace}
\usepackage{multirow}
\usepackage{lineno}
\usepackage{mathpazo}
\usepackage{avant}
\usepackage{inconsolata}
\usepackage{verbatim}
\usepackage{cancel}

\DeclareMathOperator*{\argmax}{arg\,max}
\DeclareMathOperator*{\argmin}{arg\,min}

\newcommand{\EE}{\mathbb{E}}

\newcommand{\NN}{\mathbb{N}}

\newcommand{\bfone}{\mathbf{1}}

\newcommand{\iid}{\overset{\mathrm{iid}}{\sim}}

\newcommand{\bfm}{\mathbf{m}}

\newcommand{\bfu}{\mathbf{u}}

\newcommand{\bfx}{\mathbf{x}}
\newcommand{\bfy}{\mathbf{y}}

\newcommand{\bfI}{\mathbf{I}}

\newcommand{\bfL}{\mathbf{L}}

\newcommand{\bfS}{\mathbf{S}}

\newcommand{\bfX}{\mathbf{X}}

\newcommand{\bfZ}{\mathbf{Z}}

\newcommand{\bfalpha}{\bm{\alpha}}
\newcommand{\bfbeta}{\bm{\beta}}
\newcommand{\bfgamma}{\bm{\gamma}}

\newcommand{\bfepsilon}{\bm{\epsilon}}

\newcommand{\bfxi}{\bm{\xi}}
\newcommand{\bfpsi}{\bm{\psi}}
\newcommand{\bftheta}{\bm{\theta}}

\newcommand{\bflambda}{\bm{\lambda}}
\newcommand{\bfmu}{\bm{\mu}}

\newcommand{\bfSigma}{\bm{\Sigma}}

\usepackage[ruled]{algorithm2e}
\SetKwComment{Comment}{/* }{ */}

\jyear{2022}%

\theoremstyle{thmstyleone}%
\theoremstyle{thmstyletwo}%

\theoremstyle{thmstylethree}%

\begin{document}

\title[Variational Inference for Bayesian Bridge Regression]{Variational Inference for Bayesian Bridge Regression}


\author*[1]{\fnm{Carlos Tadeu Pagani} \sur{Zanini}}\email{carloszanini@dm.ufrj.br}

\author[1]{\fnm{Helio  S.} \sur{Migon}}\email{migon@im.ufrj.br}

\author[2]{\fnm{Ronaldo} \sur{Dias}}\email{dias@unicamp.br}

\affil*[1]{\orgdiv{Departamento de M\'etodos Estat\'isticos}, \orgname{Universidade Federal do Rio de Janeiro}, \orgaddress{\city{Rio de Janeiro}, \state{RJ}, \country{Brazil}}}

\affil[2]{\orgdiv{Instituto de Matemática, Estatística e Ciência da Computação}, \orgname{Universidade Estadual de Campinas}, \orgaddress{\city{Campinas}, \state{SP}, \country{Brazil}}}


\abstract{The bridge approach for regularization of coefficients in regression models uses $\ell_{\alpha}$ norm, with $\alpha \in (0, +\infty)$, to define a penalization on large values of the regression coefficients. Particular cases include the lasso and ridge  penalizations. In Bayesian models, the penalization is enforced by a prior distribution on the coefficients. Although MCMC approaches are available for Bayesian bridge regression, they can be very slow for large datasets, specially in high dimensions. This paper develops an implementation of Automatic Differentiation Variational Inference for Bayesian inference on semi-parametric regression models with bridge penalization. The non-parametric effects of covariates are modeled by B-splines. The proposed inference procedure allows the use of small batches of data at each iteration (due to stochastic gradient based updates), therefore drastically reducing computational time in comparison with MCMC. Full Bayesian inference is preserved so joint uncertainty estimates for all model parameters are available. A simulation study shows the main properties of the proposed method and an application to a large real dataset is presented.}




\maketitle

\section{Introduction and related works}
\label{sec1}

It is not uncommon for an experimenter to be interested in understanding how covariates might explain a response variable. For this, one can assume a general non-parametric regression model:
%$$\bfy = \bfX \bfbeta + g(\bfx_{i,1},\ldots, \bfx_D) + \bfepsilon,$$ with $\bfepsilon=(\epsilon_1, \dots, \epsilon_n$ distributed according to Normal distribution $ N(\mathbf{0}, \phi \bfI)$, where $\bfy = (y_1,\ldots, y_n)$, $\bfx_d = (x_{d,1}, \dots, x_{d,n}) \; \; \mbox{and} \; \; d=1,\ldots D.$ 

\noindent $$y_i = g(x_{i1},\ldots,x_{iD}) + \epsilon_i,$$ where $\epsilon_i \sim N(0,\phi^{-1})$ for $i= 1,\ldots,n$.\\
A usual approach to estimate the surface $g$ is to consider the well-known (General) Additive Model, which briefly means that $g$ can be decomposed as
$$g(\bfx_1,\ldots, \bfx_D) = g_0 + \sum_{j=1}^{D} g_j(\bfx_j) ,$$ where $\bfx_j=(x_{1j}, \ldots, x_{nj})^{\top}$, for $j=1,\ldots, D$ and $i=1,\ldots,n$. Each univariate function $g_j$ can be written as a linear combination of basis functions $B_{j,1}, \ldots, B_{jK_j}$, such as B-splines, wavelets, Fourier basis, etc. 
That is, $g_j(\bfx_j) = \sum_{k=1}^{K_j} B_{jk}(\bfx_j) \beta_{jk}$.
%That is, $g_j(\bfx_j) = \sum_{k=1}^{K_j} B_{jk}(\bfx_j) \beta_{jk}=\bfX_j \bfbeta_j$, where $\bfX_j$ is $n\times K_j$ matrix $\bfX_j = [B_{j1}(\bfx_j), \ldots, B_{jK_j}(\bfx_j)]$ and $\bfbeta_j = (\beta_{j1}, \ldots, \beta_{jK_j})^{\top}$. In the frequentist framework the vector of coefficients $\bfbeta$ can be estimated via the backfitting algorithm. For details, see \citep{hastie1987}.

Throughout this work, the basis functions chosen to represent the univariate function $g_j$ are the well-known B-splines. Thus, the surface regression model is:
\begin{equation}
\label{modelo_B_splines}
y_i = g_0 +  \sum^D_{j=1}\sum_{k=1}^{K_j} B_{jk}(x_{ij}) \beta_{jk} + \epsilon_i
\end{equation} 

%In particular, the vector notation is $\bfg_j = \bfX_j\bfbeta_j $ and the regression model becomes:
%$$ \bfy = \bfX_0\bfbeta_0 + \sum_{j=1}^{D} \bfX_j \bfbeta_j + \bfepsilon,$$ where $ \bfepsilon \sim N(\mathbf{0}, \phi \bfI_n).$ Note that the term $\bfX \bfbeta_0$ may include the constant term $g_0.$
Regularization plays a very important role in Statistics: penalizing overcomplex models often reduces the risk of overfitting and produces better generalization to hold out data. For this, under the frequentist point of view, one have to solve the following optimization problem: 
$$ \argmin_{\bfbeta} \  \lVert \bfy - g_0 - \sum_{k=1}^{K_j} B_{jk}(\bfx_j) \beta_{jk} \rVert^2  + \sum_{j=1}^D \lambda_j \; {\mathcal{P}}(  \bfbeta_j, \alpha_j), $$ where $\lambda_j$ is the smoothing parameter and ${\mathcal{P}}(  \bfbeta_j, \alpha_j)$ is the roughness penalty term. %with $\bfbeta=(\bfbeta_1,\ldots \bfbeta_D)$.

Different types of penalties, ${\mathcal{P}}(  \bfbeta_j, \alpha) $ can be applied. For instance, ${\mathcal{P}}(\bfbeta_j, \alpha=1)= \lVert\bfbeta_j \rVert_1$, $\ell_1$-norm,
${\mathcal{P}}(  \bfbeta_j, \alpha=2)= \lVert \bfbeta_j\rVert^2_2 $, the $\ell_2$-norm squared, the smoothing splines penalty, ${\mathcal{P}}( \bfbeta_j, \alpha=2)= \int (\sum_j \bfbeta_j \ddot B_j)^2,$ and P-spline, ${\mathcal{P}}( \bfbeta_j, \alpha=2)= \sum_{j=k+1}^{D}  (\Delta^k \bfbeta_j)^2$, where $\Delta^k$ is the $k$-th order difference operator.

In particular, after writing (\ref{modelo_B_splines}) in matrix form, the semi-parametric regression model can be written as:
$$ \bfy = \bfX_0\bfbeta_0 + \sum_{j=1}^{D} \bfX_j \bfbeta_j + \bfepsilon,$$ where $ \bfepsilon \sim N(\mathbf{0}, \phi^{-1} \bfI_n)$, with $\sum_{k=1}^{K_j} B_{jk}(\bfx_j) \beta_{jk}=\bfX_j \bfbeta_j$, where $\bfX_j$ is the $n\times K_j$ matrix $\bfX_j = [B_{j1}(\bfx_j), \ldots, B_{jK_j}(\bfx_j)]$ and $\bfbeta_j = (\beta_{j1}, \ldots, \beta_{jK_j})^{\top}$.
Note that the matrix $\bfX_0 $ comprises possible parametric covariates and hence can also be viewed as Partial Splines \citep{silv:gree:1994}. In particular, if $\bfX_0= \bfone$ then $\bfX_0\bfbeta_0$ plays the role of $g_0$ from equation \eqref{modelo_B_splines} (Additive Model).
 
Penalties defined in terms of $\ell_{\alpha}$ norm in the context of Bayesian regression models include the ridge (\citealt{hoerl1970b}, \citealt{hoerl1970a}) ($\alpha = 2$), the lasso \citep{park2008} ($\alpha = 1$), elastic net \citep{li2010} (convex combination of both $\ell_1$ and $\ell_2$ norms) and bridge \citep{polson2014, mallick2018} ($\alpha > 0$), first presented by \cite{frank1993}. 
A possible bridge model for semi-parametric regression is a regularization problem defined as,
\begin{equation}
    \argmin_{\bfbeta_0,\ \bfbeta}  \frac{1}{2}\lVert(\bfy - \bfX_0 \bfbeta_0 - \sum_{j=1}^{D}\bfX_j \bfbeta_j)\Vert^2  +  \sum^D_{j=1} \lambda_j \lvert\lvert\bfbeta_j \lvert\lvert^{\alpha_j}_{\alpha_j}
    \label{eq:bridge_classical}
\end{equation}

\noindent where $\lambda_j>0$ is the penalty term that controls the strength of shrinkage over $\bfbeta_j$ and $\alpha_j > 0$ is the concavity parameter of the penalty function.

\begin{figure*}
    \centering
    \includegraphics[scale=0.4]{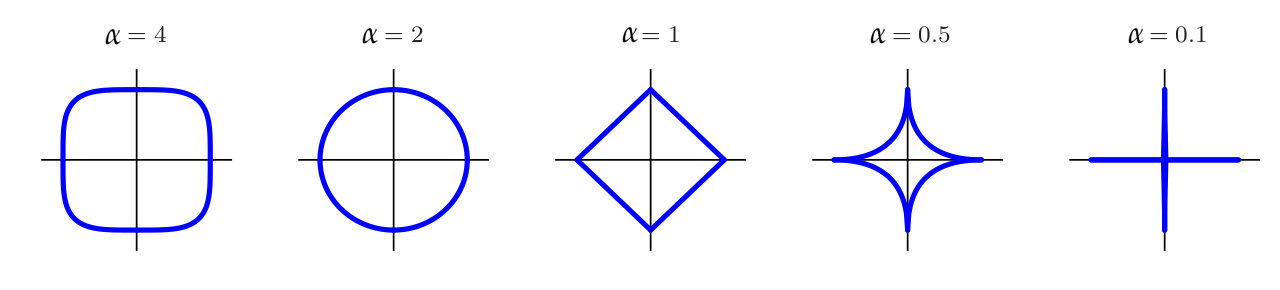}
    \caption{Constraint regions $\sum^{p_x}_{j=1} \mid\beta_j \mid^{\alpha}\leq 1$ for different values of $\alpha$. For $\alpha<1$, the constraint region is nonconvex. Figure extracted from \cite{hastie2015} (adapted).}
    \label{fig:alpha_norm_penalties}
\end{figure*}

The different choices of $\ell_{\alpha}$ norm for the penalization imply different forms of shrinking the regression coefficients towards zero. While ridge regression does not zero out coefficients, the lasso penalization is capable of producing sparse solutions to the corresponding maximum a posteriori objective, therefore working as a variable selection procedure \citep{tibshirani1996, tibshirani1997}.  From a Bayesian perspective, \cite{hastie2015}, \cite{Casella_2010} and \cite{Leng_2014}, highlight that one can interpret the term $\lVert\bfbeta_j\lVert^{\alpha_j}_{\alpha_j}$ as proportional to the negative log-prior density of $\bfbeta_j$ with the contours illustrated in Figure \ref{fig:alpha_norm_penalties} representing the contours of the prior distribution. The case $\alpha_j < 1$ implies a non-convex prior that concentrates more mass along the coordinates' axis, producing solutions with fewer nonzero coefficients and less shrinkage. 

Other forms of penalization are also relevant. The elastic net provides sparsity of representation similarly to the lasso while also encouraging a grouping effect where strongly correlated predictors tend to be in or out of the model together \citep{zou2005}. In Bayesian regression models, the horseshoe prior implies a form of variable selection that acts strongly on coefficients of small magnitude while having little influence on coefficients of large magnitude \citep{carvalho2009}. 

In regression models that include non-parametric effects, it is crucial to assign some form of penalization in order to calibrate the level of complexity of the relationships between the covariates and the response to avoid overfitting. Generalized additive models (GAM) from \cite{hastie1986} 
%multiple splines 
models the non-parametric effects of covariates on the mean of a response variable belonging to the exponential family, with one spline being used for each covariate. For common choices of spline basis functions such as B-spline, see \cite{eilers1996}.  The level of smoothness can be controlled by a penalization acting on the splines coefficients, see \cite{currie2002}. Some Bayesian approaches in this context include \cite{lang2004} and \cite{hastie2000}. In the context of bridge penalized linear regression with Gaussian response, \cite{polson2014} and \cite{mallick2018} develop full Bayesian inference through fixed dimension MCMC based on variable augmentation schemes that guarantee conjugacy for the full conditional distributions. \cite{denison1998} and \cite{dias2002} describe reversible jumps MCMC algorithms for selecting the number of knots in B-splines and P-splines. These MCMC approaches, however, are computationally very expensive when applied to large datasets. 

This work proposes an alternative to MCMC for approximate inference on Bayesian bridge semi-parametric regression with B-splines based on the Automatic Differentiation Variational Inference (ADVI) by \cite{kucukelbir2015} and \cite{kucukelbir2017}. The proposed inference algorithm jointly estimates all parameters in the model (including $\alpha_j$ and $\lambda_j$)  and drastically increases computational speed for large datasets in comparison with MCMC implementations since stochastic gradient updates involved in the underlying optimization process small batches of data at each iteration, rather than the entire dataset as required by MCMC schemes. A key point in variational inference concerns the choice of the variational family of distributions, which needs to be tractable while at the same time flexible enough to approximate the posterior distribution well. \cite{armagan2009} describe a mean-field variational inference method for Bayesian bridge regression model with approximate inference for the bridge parameter ($\alpha$). \cite{alves2021} also proposes a variational approach based on mean-field assumption for Bayesian inference in regression models with splines, however it is restricted to Lasso penalization. In contrast, the ADVI does not require the oversimplifying mean-field assumption for the variational family, as required by coordinate ascending variational inference \citep{blei2017}, producing very close approximations to the full posterior distribution, as demonstrated in the simulation study. Previous gradient based variational procedures, such as \cite{ranganath2014} or \cite{kingma2013} could be used, but they imply more restricted forms of dependence in the variational family and produce noisier estimates of the gradients when compared to ADVI. More recent approaches, such as \cite{yin2018}, allow for more flexible variational families, however we found the simpler ADVI approach to produce already very good results in terms of computational speed and posterior approximation under the semi-parametric Bayesian bridge model. 

To summarize, the main contribution of the paper is the development of a full Bayesian inference procedure based on variational inference for semi-parametric regression with bridge penalization. As strengths of the proposed methodology, we can list the following:

\begin{enumerate}
\item The proposed inference method is flexible enough to capture the dependence structure in the target posterior distribution since the variational family does not require the mean-field assumption. More specifically, for the mean-field approach, the joint variational distribution of the parameters is the product of its marginals. On the other hand, ADVI admits dependence structure given by a transformation of a multivariate normal distribution with full covariance matrix.

\item It enables full Bayesian inference on semi-parametric regression for large datasets at drastically lower computing times when compared with more traditional MCMC implementations due to batch processing of data by the stochastic gradient updates specially for large data.
\end{enumerate}

The remainder of the paper is structured as follows. The Bayesian bridge semi-parametric model is described in section \ref{sec:bayesian_bridge}. Section \ref{sec:mcmc_bridge} covers the MCMC scheme by \cite{mallick2018}, used for comparison with the proposed ADVI for Bayesian bridge semi-parametric regression defined in section \ref{sec:vb}. Section \ref{sec:simulation_study} concerns a simulation study with focus on large data. Section \ref{sec:real_data} contains an application to real data on energy charges in Brazil and, finally, section \ref{sec:conclusion} presents the conclusions and future works. 

\section{The Bayesian bridge model for semi-parametric regression}
\label{sec:bayesian_bridge}

A Bayesian bridge model for multiple covariates, can be written in the form of a hierarchical model as follows:

\begin{multline}
    (\bfy \mid \bfbeta_0, \bfbeta_1, \ldots, \bfbeta_D, \phi) \sim \\ \sim N\left(\bfX_0\bfbeta_0 + \sum^D_{j=1}\bfX_j \bfbeta_j, \ \phi^{-1}\bfI_n \right),\\
    (\beta_{jk} \mid \lambda_j, \phi, \alpha_j) \iid GG\left(0, \ \lambda^{-\frac{1}{\alpha_j}}_j\phi^{- \frac{1}{2} }, \ \alpha_j \right), \\ \ j=1, \ldots, D, \ k = 1, \ldots, K_j.
    \label{eq:GG}     
\end{multline}

\noindent where $\bfy \in \mathbb{R}^n,$ $N(\bfmu, \bfSigma)$ denotes the multivariate Gaussian distribution with mean $\bfmu$ and covariance matrix $\bfSigma,$ $GG(\mu, \sigma, \alpha)$ denotes the generalized Gaussian distribution \citep{subbotin1923} with mean $\mu$, scale $\sigma$ and shape $\alpha$. The generalized Gaussian density evaluated at $x \in \mathbb{R}$ is denoted by $$GG(x \mid \mu, \sigma, \alpha) = \frac{\alpha}{2\sigma\Gamma(\alpha^{-1})} \exp\left\{ -\left(\frac{ \mid x - \mu \mid }{\sigma} \right)^{\alpha}\right\}.$$ 

Equation \eqref{eq:GG} therefore implies the prior probability density function

\begin{multline}
p(\bfbeta_{jk} \mid \lambda_j, \phi, \alpha_j) = \\ = \frac{\alpha_j \lambda^{\frac{1}{\alpha_j}}_j \phi^{\frac{1}{2} }}{2\Gamma (\alpha^{-1}_j)} \exp\left\{-\lambda_j \left(\phi^{\frac{1}{2}}\mid \beta_j \mid \right)^{\alpha_j}\right\},
\label{eq:bayesian_lasso}
\end{multline}

\noindent so that maximizing the posterior density $p(\bfbeta \mid \bfy, \bflambda, \phi, \bfalpha, \bfbeta_0)$ over $\bfbeta$ and $\bfbeta_0$ for fixed $(\bflambda, \phi, \bfalpha)$ is equivalent to solving \eqref{eq:bridge_classical}, where $\bfbeta = (\bfbeta_1, \ldots, \bfbeta_D), \ \bflambda = (\lambda_1, \ldots, \lambda_D), \ \bfalpha = (\alpha_1, \ldots, \alpha_D)$. The particular form of dependency of $\beta_{jk}$ on $\lambda_j$ and $\phi$ expressed by equation \eqref{eq:bayesian_lasso} avoids undesired multimodality of the posterior distribution as pointed out by \cite{park2008}.

The proposed semi-parametric model splits the regressors into matrices $\bfX_j, \ j= 1, \ldots, D$ of covariates with their own penalized effects and a matrix $\bfX_0$ of covariates with unpenalized effects. In the proposed formulation, $\bfX_j$ represents a spline basis function that implies non-parametric effects to the corresponding covariates. B-splines were chosen as the set of spline basis functions to build $\bfX_j$ with the bridge penalization acting on the B-spline coefficients. Naturally, other types of spline basis functions could also be used. 

Within the Bayesian framework, it is possible to provide joint (approximate) posterior estimates for all parameters, including $\bfalpha$ and $\bflambda$ by specifying a prior distribution on them, as will be shown further in sections \ref{sec:mcmc_bridge} and \ref{sec:vb}. Therefore, joint posterior inference can be done in a single-step procedure.

\section{MCMC on Bayesian bridge regression} 
\label{sec:mcmc_bridge}

Both \cite{polson2014} and \cite{mallick2018} propose MCMC schemes for Bayesian bridge regression for fixed $\alpha_j$ based on different data augmentation schemes. The algorithm by \cite{mallick2018} was chosen for comparison with the proposed ADVI for Bayesian bridge, mostly because of the simplicity of its gamma-uniform variable augmentation in comparison with the one proposed by \cite{polson2014}. Accordingly, one can recover \eqref{eq:GG} by specifying

\begin{gather}
(u_{jk} \mid \alpha_j, \lambda_j) \sim Ga\left(\frac{1}{\alpha_j}+1, \ \lambda_j\right),  \ \ \ k = 1, \ldots, K_j,\nonumber\\
(\beta_{jk} \mid u_{jk}, \phi, \alpha_j) \sim Unif\left( -u^{\frac{1}{\alpha_{j}}}_{jk}\phi^{-\frac{1}{2}}, \ u^{\frac{1}{\alpha_j}}_{jk}\phi^{-\frac{1}{2}} \right), \nonumber \\ 
\ \ \ \ \ \ \ \ \ \ \ \ \ \ \ \ \ \ \ \ \ \ \ \ \ \ \ \ \ \ \ \ \ \ \ \ \ \ \ \ \ \ \ \ \ \ \ \ j = 1, \ldots, D.
\end{gather}

\noindent In fact,
\begin{align*}
    p(\beta_{jk} &\mid \lambda_j, \alpha_j, \phi) = \\
    &=\int^{\infty}_0 p(\beta_{jk} \mid u_{jk}, \phi, \alpha_j) p(u_{jk} \mid \alpha_j, \lambda_j) du_{jk}\\
    &\propto \exp\{-\lambda_j \phi^{\frac{\alpha_j}{2}}\mid \beta_{jk} \mid^{\alpha_j}\}.
\end{align*}

The model specification is completed by assigning independent priors $\phi \sim Ga(a_{\phi}, b_{\phi}),$ $\bfbeta_0 \sim N(\bfmu_{0}, \Sigma_{0}), \lambda_j \sim Ga(a_{\lambda}, b_{\lambda}),$ where $Ga(a,b)$ denotes the Gamma distribution with mean $a/b$ and variance $a/b^2$. The prior distribution on $\alpha_j$ was defined by taking $\alpha_j = 2.5\eta_j$ where $\eta_j\sim Beta(a_{\eta}, b_{\eta} )$. The upper limit for $\alpha_j$ in this work is fixed arbitrarily at 2.5 so that the lasso and the ridge penalizations can be approximately represented as special cases of bridge regression.

Gibbs sampler can be easily implemented for posterior inference on the Bayesian bridge regression model. Due to conjugacy results, the full conditional distributions of all parameters, except $\alpha_j$, are analytically available. Details concerning the MCMC implementation are described in the appendix \ref{sec:MCMC_details}. In section \ref{sec:gibbs}, the Gibbs sampler approach considering $\alpha_j$ fixed is described. To include $\alpha_j$ in the sampling scheme, one can carry out a Metropolis-Hastings step with a transformed random walk proposal as described in section \ref{sec:MH_alpha} (when the auxiliary vector $\bfu$ is marginalized) or in section \ref{append:MH} when $\bfu$ is not marginalized). It was found that the marginalized random walk proposal produce well mixing Markov chains without much effort to tune the proposal variance. Naturally, other Metropolis-Hastings proposals could be used.

\section{Variational Inference}
\label{sec:vb}

The variational approach searches among a predefined variational family $\mathcal{Q} = \{q_{\bfpsi}(\bftheta): \ \bfpsi\in \mathcal{V}\}$ (where its members are densities indexed by the variational parameter $\bfpsi$) for the density $q_{\bfpsi^*}(\bftheta)$ that best approximates the posterior $p(\bftheta \mid \bfy)$ in terms of Kullback-Leibler divergence. For the proposed Bayesian semi-parametric bridge regression, $\bftheta = (\bfbeta, \bfgamma, \phi, \lambda, \alpha)^{\top} \in \mathbb{R}^{p_X} \times \mathbb{R}^{p_Z} \times (0,\ 1)^2 \times (0, \ 2.5).$

Therefore, the variational objective is $$ \argmin_{\bfpsi\in \mathcal{V}} KL\left( q_{\bfpsi}(\bftheta) \mid \mid p( \bftheta\mid \bfy) \right),$$ which is equivalent to maximize the evidence lower bound (ELBO), i.e. $\argmax_{\bfpsi\in \mathcal{V}} ELBO(\bfy, \bfpsi),$ where 

\begin{align*}
E&LBO(\bfy, \bfpsi):=\mathbb{E}_{q_{\bfpsi}}[\log p(\bfy, \bftheta) - \log q_{\bfpsi}(\bftheta)]\\
&=\int \left[ \log p(\bfy, \bftheta) - \log q_{\bfpsi}(\bftheta)\right]q_{\bfpsi}(\bftheta) \ d\bftheta.
\end{align*}

One way to maximize the ELBO is to calculate its gradient $\nabla_{\bfpsi}ELBO(\bfy, \bfpsi)$ and use it in a stochastic gradient ascend based algorithm, such as Adam (\cite{adam}), Adagrad \citep{duchi2011}, AdaDelta \citep{zeiler2012} and others. The issue with such gradient ascending methods is evaluating the expectation in the ELBO, which is often intractable. In this case, a common solution involves rewriting the ELBO in a way that the gradient operator can switch order with the expectation so the resulting $\nabla_{\bfpsi}ELBO(\bfy, \bfpsi)$ is an expected value with respect to the variational distribution. The resulting expectation can then be estimated via Monte Carlo methods. 

This work implements ADVI (section \ref{sec:advi}) to the semi-parametric Bayesian bridge regression described in section \ref{sec:bayesian_bridge}. The variational approaches by \cite{ranganath2014} and by \cite{kingma2013} are briefly presented for comparison and contextualization. A reader interested in variational inference for semi-parametric regression models may see \citep{luts2015}, \citep{Menictas2015}, \citep{Ong2017} and \citep{Wand2017} to mention a few.

\subsection{Score method}

\cite{ranganath2014} introduced the black box variational inference (BBVI) in which the score method (also known as "log trick") allows the gradient of the ELBO to be written as an expectation with repect to the variational distribution. The resulting expectation can be estimated by Monte Carlo as follows

\begin{multline}
    \nabla_{\bfpsi} ELBO(\bfy, \bfpsi)= \\
    =\int \nabla_{\bfpsi} \left\{[\log p(\bfy \mid \bftheta) + \log p(\bftheta) + \right.\\ 
    \ \ \ \ \ \ \ \ \ \ \ \ \ \ \ \ \ \ \ \ \ \ \ \ \ \ \ \ \ \ \ \ \ \ \ \ \ \ \ \ \ \ \ \ \ \left. - \log q_{\bfpsi}(\bftheta)] q_{\bfpsi}(\bftheta) \right\}d\bftheta \\
    =\mathbb{E}_{q_{\bfpsi}}[( \log p(\bfy \mid \bftheta) + \log p(\bftheta) - \log q_{\bfpsi}(\bftheta) )\times \\ 
    \ \ \ \ \ \ \ \ \ \ \ \ \ \ \ \ \ \ \ \ \ \ \ \ \ \ \ \ \ \ \ \ \ \ \ \ \ \ \ \ \ \ \ \ \ \ \ \ \ \ \ \ \ \ \  \times \nabla_{\bfpsi} \log q_{\bfpsi} (\bftheta)] \\
    \approx \frac{1}{M} \sum^M_{m=1} \left[\log p(\bfy \mid \bftheta^{(m)}) + \log p(\bftheta^{(m)}) + \right. \\ 
    \left. - \log q_{\bfpsi}(\bftheta^{(m)}) \right] \nabla_{\bfpsi} \log q_{\bfpsi} (\bftheta^{(m)}), 
    \label{eq:MC_BBVI}
\end{multline}

\noindent where \ \ $\bftheta^{(m)} \iid q_{\bfpsi}(\bftheta), \ m = 1 \ldots, M.$

Evaluating \eqref{eq:MC_BBVI} requires (i) the prior, likelihood and variational densities to be analytically available and (ii) the ability to draw samples from the variational distribution $q_{\bfpsi}(\bftheta)$. However, in many applications, the Monte Carlo estimate from \eqref{eq:MC_BBVI} has high variance \citep{paisley2012}, even when control variates and Rao-Blackwellization are used to reduce variance. It is simple to mitigate the mean field hypothesis in the BBVI method to some extent by grouping blocks of components within $\bftheta$ and specifying a dependent multivariate variational distribution within each block. However specifying a single dependent distribution for the entire $\bftheta$ is typically hard, specially when its components lie in different subspaces of $\mathbb{R}$ (so a multivariate Gaussian or Student-t would be inappropriate choices for $q_{\bfpsi}$). We found it crucial to account for full dependence structure in the variational family for the case of semi-parametric Bayesian bridge model. In Appendix \ref{append:bbvi}, we provide the specifics on how one can apply the BBVI algorithm to the Bayesian bridge model.

\subsection{Reparameterization gradient}

An alternative way to derive Monte Carlo estimates for the gradient of the ELBO is described in \cite{kingma2013}. The authors present the reparameterization trick, which assumes that the parameter vector $\bftheta$ can be analytically written as $\bftheta = T_{\bfpsi}(\bfepsilon)$ where $T_{\bfpsi}$ is a differentiable deterministic transformation involving the variational parameters $\bfpsi$ and $\bfepsilon$ is a random noise required to have a closed form distribution $q^*(\bfepsilon)$ that does not depend on $\bfpsi$ and is easy to sample from. For example, if $\bftheta \sim N(\bfm, \bfS),$ then we can write $\bfbeta = T_{\bfm, \bfS}(\bfepsilon) = \bfm + \bfL\bfepsilon,$ with $\bfL$ being the Choleskey decomposition of the covariance matrix $\bfS$ (ie, $\bfL \bfL^{\top} = \bfS$) and $\bfepsilon \sim N(\boldsymbol{0}, \bfI).$ Another example of reparameterization is the log-Normal distribution: if $\theta \sim \log N(m, s^2),$ one can write $\theta = T_{m,s}(\epsilon) = \exp\{m + s \times \epsilon \}$ with $\epsilon \sim N(0, 1).$

The reparameterization of $\bftheta$ allows one to switch the order of the gradient with the expectation when deriving the updating equations for optimizing the ELBO. The gradient of the ELBO under reparameterization becomes 
\begin{multline*}
\nabla_{\bfpsi} ELBO(\bfpsi) = \\ =\EE_{\bfepsilon \sim q^*}\left[\nabla_{\bfpsi} \log p(\bfy, T_{\bfpsi}(\bfepsilon))\right] + \\ - \EE_{\bftheta \sim q_{\bfpsi}(\bftheta)}[\nabla_{\bfpsi}\log q_{\bfpsi}(\bftheta)].
\end{multline*}

If a random sample \ $\bfepsilon^{(1)}, \ldots, \bfepsilon^{(M)}$ is drawn from $q^*(\bfepsilon),$ the Monte Carlo estimate for the gradient is obtained as

\begin{multline}
\widetilde{\nabla}_{\bfpsi} ELBO(\bfpsi) = \\ \frac{1}{M}\sum^{M}_{m=1}\nabla_{\bfpsi} \log p(\bfy, T_{\bfpsi}(\bfepsilon^{(m)})) \\ - \frac{1}{M}\sum^{M}_{m=1}\nabla_{\bfpsi}\log q_{\bfpsi}(T_{\bfpsi}(\bfepsilon^{(m)})).
\label{eq:grad_elbo_repar}
\end{multline}

The variance of the Monte Carlo estimates for the reparameterization gradients tend to exhibit lower variance than the BBVI Monte Carlo estimates, but the difficulty of assuming a full dependence structure on the variational family for different entries of $\bftheta$ persists. In Appendix B an implementation of the reparameterization method for variational inference on the proposed semi-parametric Bayesian bridge regression model is briefly described.

\subsection{Automatic differentiation variational inference}
\label{sec:advi}

The implementation of ADVI method from \cite{kucukelbir2015} and \cite{kucukelbir2017} to the proposed semi-parametric Bayesian bridge model is briefly described in this section. The ADVI shares similarities with the reparameterization method of \cite{kingma2013} since it also makes use of reparameterization in order to write the gradient of the ELBO as an expectation to be approximated by Monte Carlo. The key distinction is that the method works in a transformed parameter space that is (ideally) suitable to be modeled as a multivariate normal distribution. 

Suppose the original parameters $\bftheta$ vary on a subset $\Theta \subset \mathbb{R}^d$ with $\Theta \neq \mathbb{R}^d.$ This happens for example when one or more entries of $\bftheta$ lie in constrained subsets of $\mathbb{R},$ say $\mathbb{R}^+$ or the interval $(0,1)$ for instance. We consider the transformed parameter vector $\bfxi = T(\bftheta)$ where $T:\Theta \rightarrow \mathbb{R}^d$ is a is diffeomorphism map (differentiable and invertible transformation) such that $\bfxi$ lies in $\mathbb{R}^d,$ with no restrictions. For the mean-field approach, the joint variational distribution of the parameters is the product of its marginals: $q_{\bfpsi}(\bftheta) = \prod^K_{j=1} q_{\psi_j}(\theta_j)$ for $\bftheta = (\theta_1, \ldots, \theta_K)$. On the other hand, ADVI admits dependence structure given by $\bftheta = T^{-1}(\bfxi)$ where $\bfxi \sim N(\bfm, \bfSigma)$ with full covariance matrix $\bfSigma$.

In the case of the Bayesian bridge model, we have the original parameters $\bftheta = (\bfbeta_0, \bfbeta_1, \ldots, \bfbeta_D, \phi, \lambda_1, \ldots, \lambda_D, \alpha_1, \ldots, \alpha_D)^{\top} \in \mathbb{R}^{K_0+K_1 + \ldots + K_D} \times (0, \ +\infty)^{D+1} \times (0, \ 2.5)^D$. A possible choice for $T$ is \begin{multline*}T(\bftheta) = (\bfbeta_0, \bfbeta_1, \ldots, \bfbeta_D, \log\phi, \log\lambda_1, \ldots, \log\lambda_D, \\ \log\frac{\alpha_1}{2.5 - \alpha_1}, \ldots, \log\frac{\alpha_D}{2.5 - \alpha_D})^{\top}.\end{multline*} In the remainder of this section, we will denote $\bfxi = (\bfxi_{\bfbeta_0}, \bfxi_{\bfbeta_1}, \ldots, \bfxi_{\bfbeta_d}, \xi_{\phi}, \xi_{\lambda_1}, \ldots, \xi_{\lambda_D}, \xi_{\alpha_1}, \ldots, \xi_{\alpha_D})^{\top},$ where $\bfxi_{\bfbeta_0} = \bfbeta_0, \ \bfxi_{\bfbeta_j} = \bfbeta_j, \ \xi_{\phi} = \log{\phi}, \ \xi_{\lambda_j} = \log{\lambda_j}, \ \xi_{\alpha_j} = \log\frac{\alpha_j}{2.5 -\alpha_j}, $ for $j = 1, \ldots, D$. The joint distribution $p(\bfy, \bftheta)$ is defined as in the right hand side of equation \eqref{eq:gibbs_joint_unmarginalized} with likelihood $(\bfy \vert \bfbeta_0, \bfbeta_1, \ldots, \bfbeta_D, \phi) \sim N(\sum^D_{j=0}\bfX_j \bfbeta_j, \phi^{-1}\bfI_n)$ and prior  $p(\bftheta)=p(\bfbeta_0)p(\phi)\prod^D_{j=1}p(\bfbeta_j \vert \lambda_j, \phi, \alpha_j)p(\lambda_j)p(\alpha_j),$ $(\beta_j \vert \lambda_j, \phi, \alpha_j) \iid GG(0, \lambda_j^{-\frac{1}{\alpha_j}}\phi^{- \frac{1}{2} }, \alpha_j)$, $\phi \sim Ga(a_{\phi}, b_{\phi}),$ $\bfbeta_0 \sim N(\bfmu_{0}, \Sigma_{0}), \lambda_j \sim Ga(a_{\lambda}, b_{\lambda})$ with $\alpha_j = 2.5\eta_j$ where $\eta_j\sim Beta(a_{\eta}, b_{\eta} )$ for $j = 1, \ldots, D.$

The ADVI method redefines the joint model density in terms of the joint distribution of $\bfy$ and the transformed parameter $\bfxi$, here denoted as $\tilde{p}(\bfy, \bfxi)$ to distinguish it from the original joint density $p(\bfy, \bftheta)$. It follows that 
\begin{align*}
\tilde{p}(\bfy, \bfxi) &= p(\bfy, \bftheta) \bigg\rvert_{\bftheta = T^{-1}(\bfxi)}\times \mid J_{T^{-1}(\bfxi)}\mid \\ &= p(\bfy, T^{-1}(\bfxi)) \times \mid J_{T^{-1}(\bfxi)} \mid,
\end{align*}
where $J_{T^{-1}(\bfxi)}$ represents the Jacobian of the inverse transformation $T^{-1}: \mathbb{R}^d \rightarrow \Theta,$

\begin{multline*}
 T^{-1}(\bfxi)  = \left(\bfxi_{\bfbeta_0}, \bfxi_{\bfbeta_1}, \ldots, \bfxi_{\bfbeta_D}, e^{\xi_{\phi}}, e^{\xi_{\lambda_1}}, \ldots, e^{\xi_{\lambda_D}}, \right.\\ \left.\frac{2.5}{1 + e^{-\xi{\alpha_1}}}, \ldots, \frac{2.5}{1 + e^{-\xi{\alpha_D}}}\right)^{\top}
\end{multline*}

\noindent which in this case is $\vert J_{T^{-1}(\bfxi)}\vert = 2.5^D e^{\xi_{\phi} } \prod^D_{j=1}e^{\xi_{\lambda_j} } e^{-\xi_{\alpha_j} } / (1 + e^{-\xi_{\alpha_j} })^2$. A multivariate Gaussian variational distribution $q(\bfxi \mid \bfL, \ \bfm) = N( \bfxi \mid \bfm, \ \bfL\bfL^{\top})$ is specified for \ $\bfxi$ and the variational parameters are $\bfpsi = (\bfm, \bfL)$. By reparameterizing $\bfxi = \bfm + \bfL \ \bfepsilon,$ the ADVI method enables calculation of the gradient of the ELBO as an expectation, which can be approximated via Monte Carlo: \begin{multline}
\nabla_{\bfpsi}ELBO(\bfy, \bfpsi) =  \\ 
=\nabla_{\bfpsi}\mathbb{E}_{q_{\bfpsi}(\bfxi)}\left[ \log \tilde{p}(\bfy, \bfxi) - \log q_{\bfpsi}(\bfxi)\right] \\
= \nabla_{\bfpsi}\mathbb{E}_{\bfepsilon \sim N(\bf0, \bfI)}\left[ \log p(\bfy, T^{-1}(\bfxi) ) +  \right. \\ 
\left. +\log \mid J_{T^{-1}(\bfxi)} \mid - \log N(\bfxi ; \ \bfm, \bfL\bfL^{\top}) \bigg\rvert_{\bfxi = \bfm + \bfL\bfepsilon}\right]\\
= \mathbb{E}_{\bfepsilon \sim N(\bf0, \bfI)}\left\{\nabla_{\bfpsi}\left[  \log p(\bfy, T^{-1}(\bfxi) ) + \log \mid J_{T^{-1}(\bfxi)}\mid \right. \right. \\ 
\left. \left. - \log N(\bfxi ; \ \bfm, \bfL\bfL^{\top}) \bigg\rvert_{\bfxi = \bfm + \bfL\bfepsilon}\right] \right\}\\
\approx \frac{1}{M} \sum^M_{\ell=1} \nabla_{\bfpsi}\left[  \log p(\bfy, T^{-1}(\bfxi) ) + \log \mid J_{T^{-1}(\bfxi)}\mid \right. \\ \left. - \log N(\bfxi ; \ \bfm, \bfL\bfL^{\top}) \bigg\rvert_{\bfxi = \bfm + \bfL\bfepsilon^{(\ell)}}\right] 
\label{eq:elbo_advi_full_gradient} 
\end{multline}

\noindent where $\bfepsilon^{(\ell)} \iid N(\bf0, \bfI)$, \ $\ell=1, \ldots, M$.

One can also easily compute the stochastic gradient approximation of \eqref{eq:elbo_advi_full_gradient} for a random minibatch $\tilde{\bfy} = \left(y_{i_1}, \ldots, y_{i_K}\right)$ where $i_1, \ldots, i_K$ represent a random subset of size $K$ from $\{1, \ldots, n\}$. There are many ways in which one can draw random minibatches $\tilde{\bfy}$, as long as the Monte Carlo estimate for the gradient based on $\tilde{\bfy}$ is unbiased for the full gradient based on $\bfy$. In this work, we randomly permute the indexes of $\bfy$ at the beginning of each epoch and pick blocks of $M$ consecutive observations to form the random batches at each iteration. The Monte Carlo estimate for stochastic gradient of the ELBO becomes

\begin{multline}
\widetilde{\nabla}_{\bfpsi} ELBO(\tilde{\bfy},\bfpsi) \\
= \frac{n}{KM} \sum^M_{\ell=1} \left[ \nabla_{\bfpsi} \log p(\tilde{\bfy} \mid T^{-1}(\bfxi) ) \bigg\rvert_{\bfxi = \bfm + \bfL\bfepsilon^{(\ell)}}\right] \ + \ \\ + \frac{1}{M} \sum^M_{\ell=1} \nabla_{\bfpsi}\left[ \log p(T^{-1}(\bfxi) ) + \log \mid J_{T^{-1}(\bfxi)} \mid \right. \\
\left. - \log N(\bfxi ; \ \bfm, \bfL\bfL^{\top}) \bigg\rvert_{\bfxi = \bfm + \bfL\bfepsilon^{(\ell)}}\right].
\label{eq:elbo_advi_stochastic_gradient}
\end{multline}

In equation \eqref{eq:elbo_advi_stochastic_gradient}, the terms $p(\tilde{\bfy} \mid T^{-1}(\bfxi) )$ and $p(T^{-1}(\bfxi) )$ denote the densities $p(\tilde{\bfy} \mid \bftheta)$ and $p(\bftheta)$ evaluated at $\bftheta = T^{-1}(\bfxi).$ The multiplicative term $\frac{n}{K}$ rescales the minibatch gradient to make it an unbiased estimate for the full posterior $p(\bftheta \mid \bfy)$ based on $n$ observations rather than the batch size posterior, which is based on $M$ data points only. Algorithm \ref{alg:proposed_advi} summarizes the steps of the proposed ADVI algorithm for semi parametric Bayesian bridge. The expression in equation \eqref{eq:elbo_advi_stochastic_gradient} is central to Algorithm \ref{alg:proposed_advi} so its details are fully developed in Appendix \ref{sec:ADVI_details}.
The update equation in line 5 corresponds to the basic stochastic gradient ascent, although any gradient based method such as Adam, Adagrad and others could be used.

\begin{algorithm}
\caption{ADVI algorithm for semi parametric Bayesian bridge}\label{alg:proposed_advi}
\SetAlgoLined
\SetKwInOut{Input}{Input}
\SetKwInOut{Output}{Output}
\SetKwInOut{Algorithm}{Algorithm}
\Input{\\
\noindent Data: $\bfy, \bfX, \bfZ$, \\
\noindent Learning rate: \ $\delta\in (0,1)$, \\
\noindent Batch size: \ $K \in \NN$\\
\noindent Number of Monte Carlo samples: \ $M\in \NN,$ \\
\noindent Number of iterations: \ $I\in \NN$,\\
\noindent Number of samples from $q_{\bfpsi}(\bftheta)$: $S \in \NN$,\\
\noindent Batch size: $B \in \NN.$} 

{\color{white} .}

{\color{white} .}

\nl \For{ $\mbox{iter} \in \{1, \ldots, I\}$}{
\nl Pick a random batch $\tilde{\bfy} = (\tilde{y}_{i_1}, \dots, \tilde{y}_{i_B})$ from $\bfy = (y_1, \ldots, y_n).$

\nl Sample $\bfepsilon^{(\ell)} \iid N(\bm{0}, \bfI), \ \ell \in \{1, \ldots, M\}.$

\nl Evaluate $\widetilde{\nabla}_{\bfpsi} ELBO(\tilde{\bfy},\bfpsi)$ from equation \eqref{eq:elbo_advi_stochastic_gradient} and Appendix \ref{sec:ADVI_details}.

\nl Gradient ascent iteration: $$\bfpsi \gets \bfpsi + \delta \times \widetilde{\nabla}_{\bfpsi} ELBO(\tilde{\bfy},\bfpsi).$$
}

\nl Sample $\bfepsilon^{(s)} \iid N(\bm{0}, \bfI), \ \ s = 1, \ldots, S.$ 

\nl Generate samples $\bftheta^{(s)} \sim q_{\bfpsi}(\bftheta)$ by taking $\bftheta^{(s)} = T^{-1}(\bfm + \bfL \bfepsilon^{(s)}), \ \ s = 1, \ldots,S.$\\

\Output{ \\ 
\noindent $\bfpsi = (\bfm, \bfL)$ 

\noindent $\bftheta^{(s)} = T^{-1}(\bfm + \bfL \bfepsilon^{(s)}), \ \ s = 1, \ldots, S.$}
\end{algorithm}

The ADVI method was found to suit well to the Bayesian bridge semi-parametric proposed model, both in terms of computational speed and also in terms of quality of posterior approximation, due to the full dependence structure incorporated in the variational family. These findings are documented in the simulation study and in the application to real data presented in sections \ref{sec:simulation_study} and \ref{sec:real_data}.

\section{Numerical experiments}
\label{sec:simulation_study}

In this section, posterior estimates under the MCMC approach based on \cite{mallick2018} as described in section \ref{sec:mcmc_bridge} and the proposed ADVI for semi-parametric Bayesian bridge, as described in section \ref{sec:advi} are compared under different scenarios, focusing on large datasets. We also compare results with the Bayesian formulations for generalized additive models provided by the R packages: mgcv (\citealt{wood2012, wood2017}), brms \citep{brms}, and INLA \citep{rue2009}. Finally, we exemplify the proposed procedure in a synthetic dataset with multiple covariates. 

\subsection{Scenario 1: small datasets}
\label{sec:simulation_1}

\begin{figure*}
    \centering
    \includegraphics[scale=0.3]{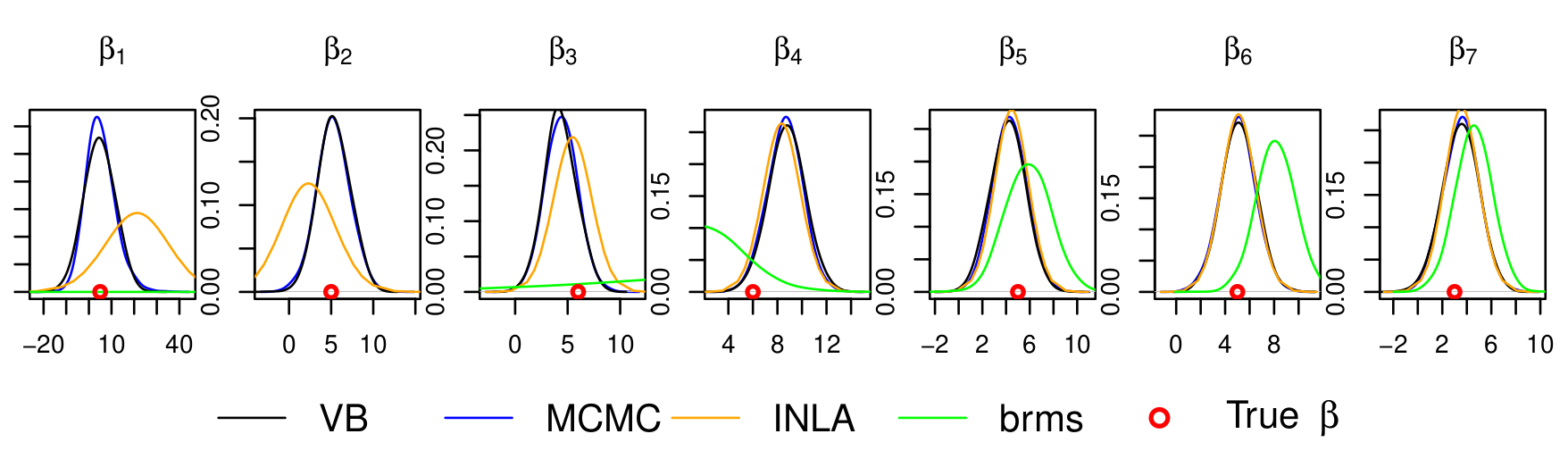}
    \caption{Marginal posterior distributions for $\beta_k, \ k=1, \ \ldots, 7$ based on the first replicate of the simulated data. Red dots denote the true values of the parameters. The marginal densities for the remaining coefficients can be seen in Figure \ref{fig:marginals_vb_mcmc_APP}. }
    \label{fig:marginals_vb_mcmc}
\end{figure*}

In this scenario, multiple small datasets are simulated and used to draw comparisons between the proposed ADVI for semi-parametric regression as described in Section \ref{sec:advi} and the MCMC scheme described in \ref{sec:mcmc_bridge}. One hundred datasets were simulated, with $n = 100$ observations each, according to the model defined by $y_i \stackrel{ind}{\sim} N( f_{\bfbeta}(x_i), \ \sigma^2), \ i = 1, \ldots, n,$ where $f_{\bfbeta}(x)$ represents the non-parametric effects of the covariate $x$, modeled by a B-splines with coefficients $\bfbeta.$ The observational variance parameter $\sigma^2$ was fixed at 1 when simulating the data and the nodes for the B-spline basis were regularly spaced from -0.066 to 1.066 with 0.033 units of consecutive distance, which implied 34 B-spline coefficients. We also fixed the covariates $x_i, \ i = 1, \ldots, 100$ on a regular grid over the interval (0, 1). The B-spline coefficients $\beta_k, \ k=1, \ldots, 34$ were sampled as follows: $\beta_1, \ldots, \beta_{10} \stackrel{iid}{\sim} Nt(5, 2),$ $\beta_{16}, \ldots, \beta_{25} \stackrel{iid}{\sim} Nt(10, 2),$ $\beta_{31}, \ldots, \beta_{34} \stackrel{iid}{\sim} Nt(4, 0.25)$ and $\beta_k = 0 \ \forall k\in\{11, \ldots, 15\}\cup\{26, \ldots, 30\}$ and the values were truncated to the nearest integers to generate the splines. The true simulated values for $\beta_k, \ k \in \{1, \ldots, 34\}$ can be seen in Figure \ref{fig:marginals_vb_mcmc} and are shared for all 100 simulated datasets, which implies only one underlying B-splines curve, along which all the replicas for $\bfy = (y_1, \ldots, y_{n})$ are simulated. Since the B-splines do not include intercept, the parameter $
\bfgamma$ does not need to be included in the model.

\begin{figure}[H]
    \centering
    \includegraphics[scale=0.22]{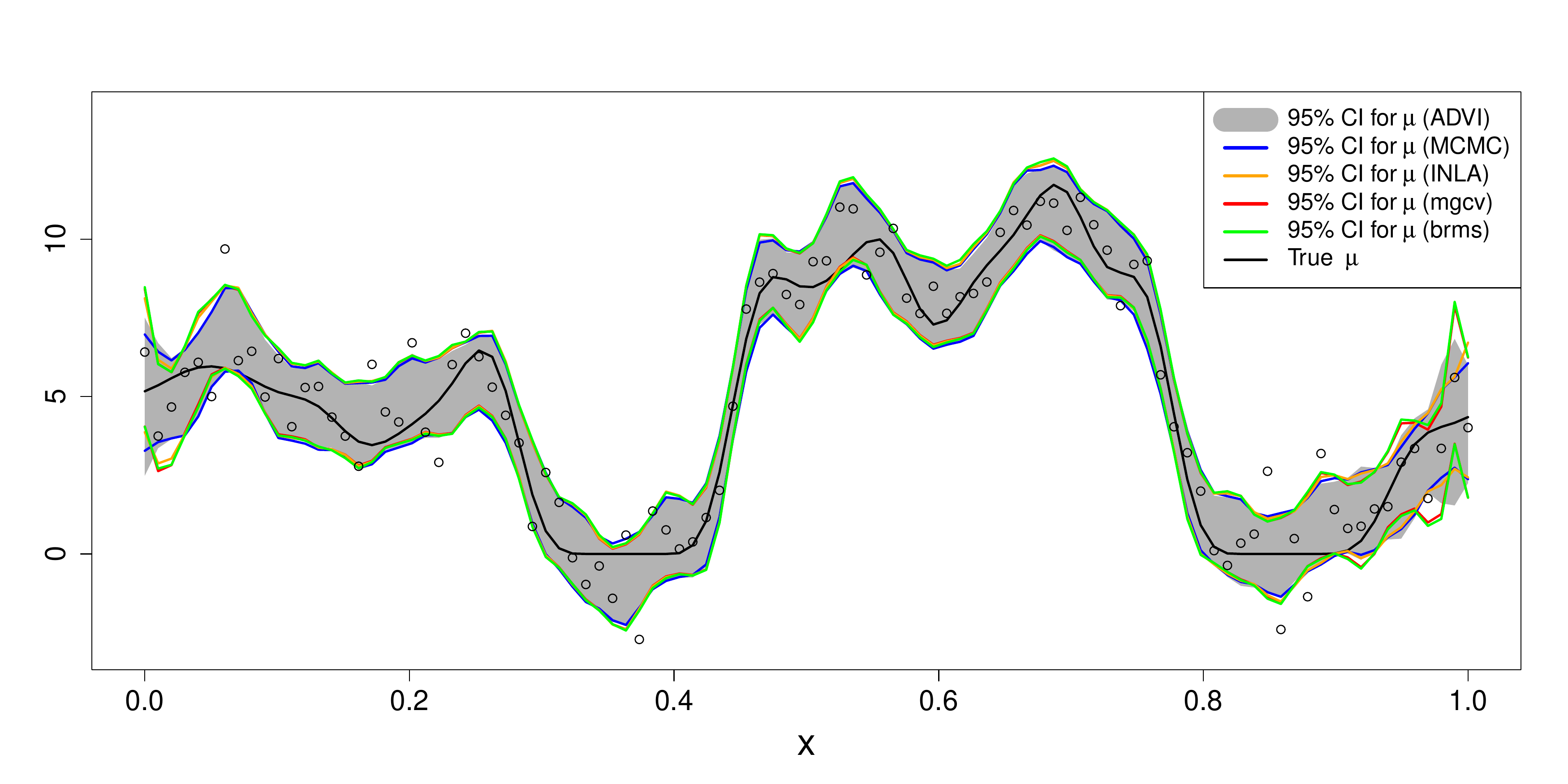}
    \caption{ADVI and MCMC posterior confidence bands for the true B-spline curve $\mu = f_{\bfbeta}(x)$ simulated in the first replica. Only one curve is shown for the point estimates of $\mu$ for ease of visualization, since both methods produced virtually indistinguishable curves.}
    \label{fig:conf_bands_first_replica}
\end{figure}

The prior hyperparameters specified for Bayesian inference under both MCMC and Variational Bayes (VB) approach via ADVI were fixed as $a_{\alpha} = a_{\lambda} = a_{\phi} = b_{\alpha} = b_{\lambda} = b_{\phi} = 1,$ therefore representing vague prior knowledge about $\alpha, \ \lambda$ and $\phi$.  Figure \ref{fig:conf_bands_first_replica} shows the posterior inference for $\mu(x) = f_{\beta}(x), \ x \in (0, \ 1)$ based on the first replica of the simulated data. MCMC and VB produce very similar posterior estimates for $\mu(x)$, with noticeable differences only at the begining and end of the series. The same holds for the chosen alternative inference methods: Integrated Nested Laplace Approximation (INLA) by \cite{rue2009}, the brms R package by \cite{brms} (which implements MCMC within Stan) and the smooth.spline R function.

\begin{figure}[H]
    \centering
    \includegraphics[scale=0.22]{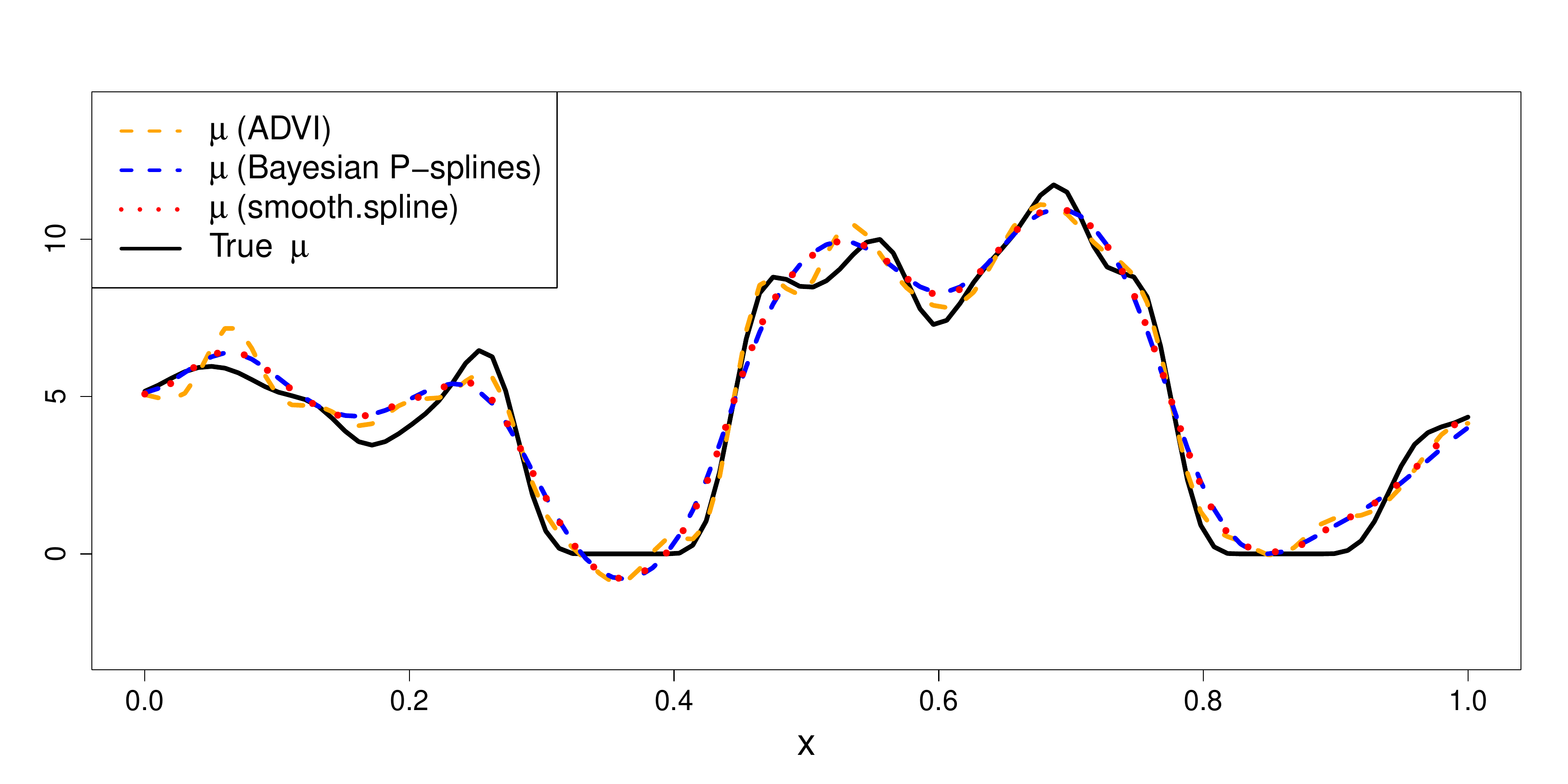}
    \caption{ Posterior estimates for the true B-spline curve $\mu = f_{\bfbeta}(x)$ simulated in the first replica obtained by ADVI and Bayesian P-splines (MCMC). The classical point estimate produced by the R function smooth.spline is also shown.} 
    \label{fig:comaprison_vb_pslines_smooth_splines}
\end{figure}

Next, Figure \ref{fig:comaprison_vb_pslines_smooth_splines} compares the ADVI with bridge penalization on the spline basis coefficients as in equation \eqref{eq:bridge_classical}, to its analogous Bayesian p-splines formulation by \cite{lang2004} (MCMC) and to the off-the-shelf smooth.spline R function. The objective is to compare the 3 distinct forms of penalization: 1) directly on the coefficients (proposed ADVI), 2) on the second order differences on consecutive coefficients (P-splines) and under no penalization (smooth.spline). The fitted curves are very similar, with the proposed ADVI for Bayesian bridge penalization being slightly less smooth than the other two approaches. Furthermore, the observed mean absolute prediction error when estimating $\mu=f_{\beta}(x)$ was lower when using the ADVI (0.4263) when compared to Bayesian P-splines (0.5648) and smoothing splines (0.5296).

\begin{figure}[H]
    \centering
    \includegraphics[scale=0.4]{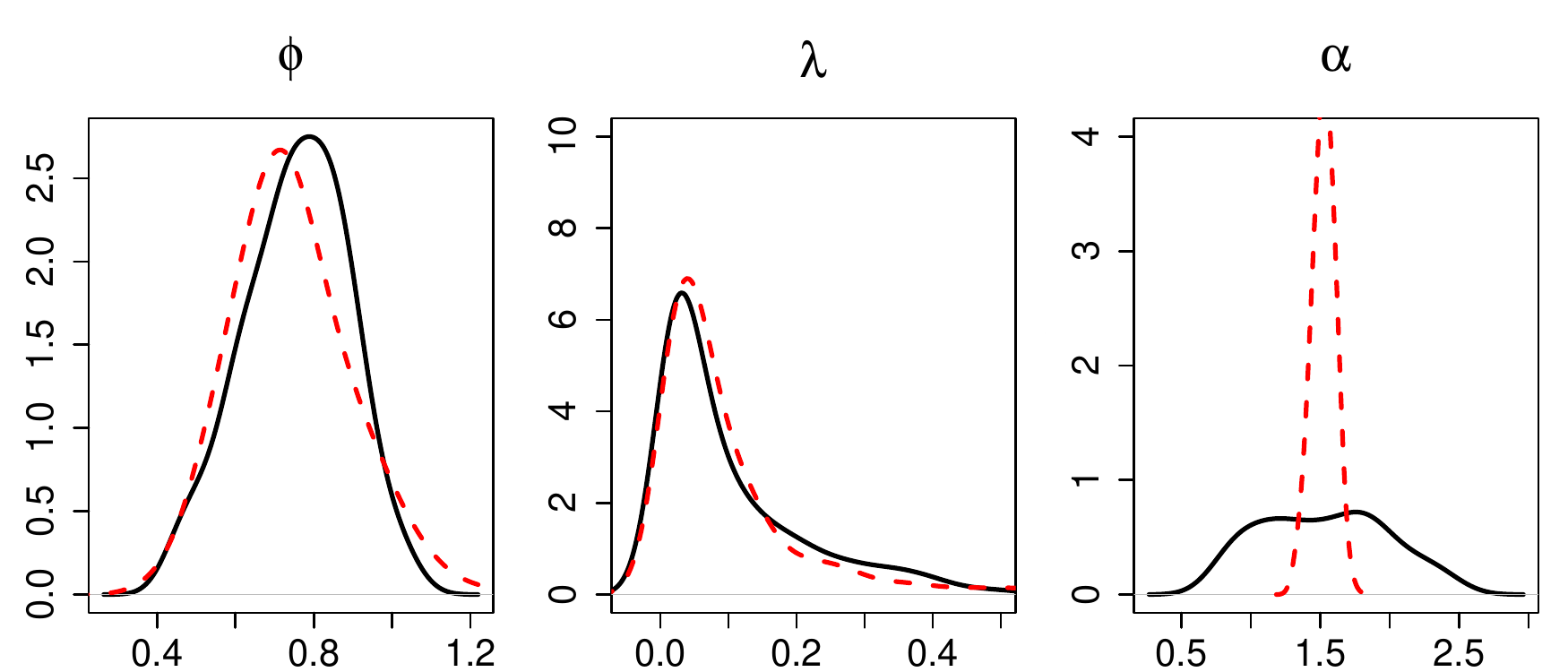}
    \caption{Marginal posterior distributions for $\phi$, $\lambda$ and $\alpha$ based on the first replicate of the simulated data. Red dashed curves represent VB and black curves represent MCMC marginal posterior approximations.}
    \label{fig:lambda_alpha_phi_vb_mcmc}
\end{figure}

\begin{figure}[H]
    \centering
    \includegraphics[scale=0.2]{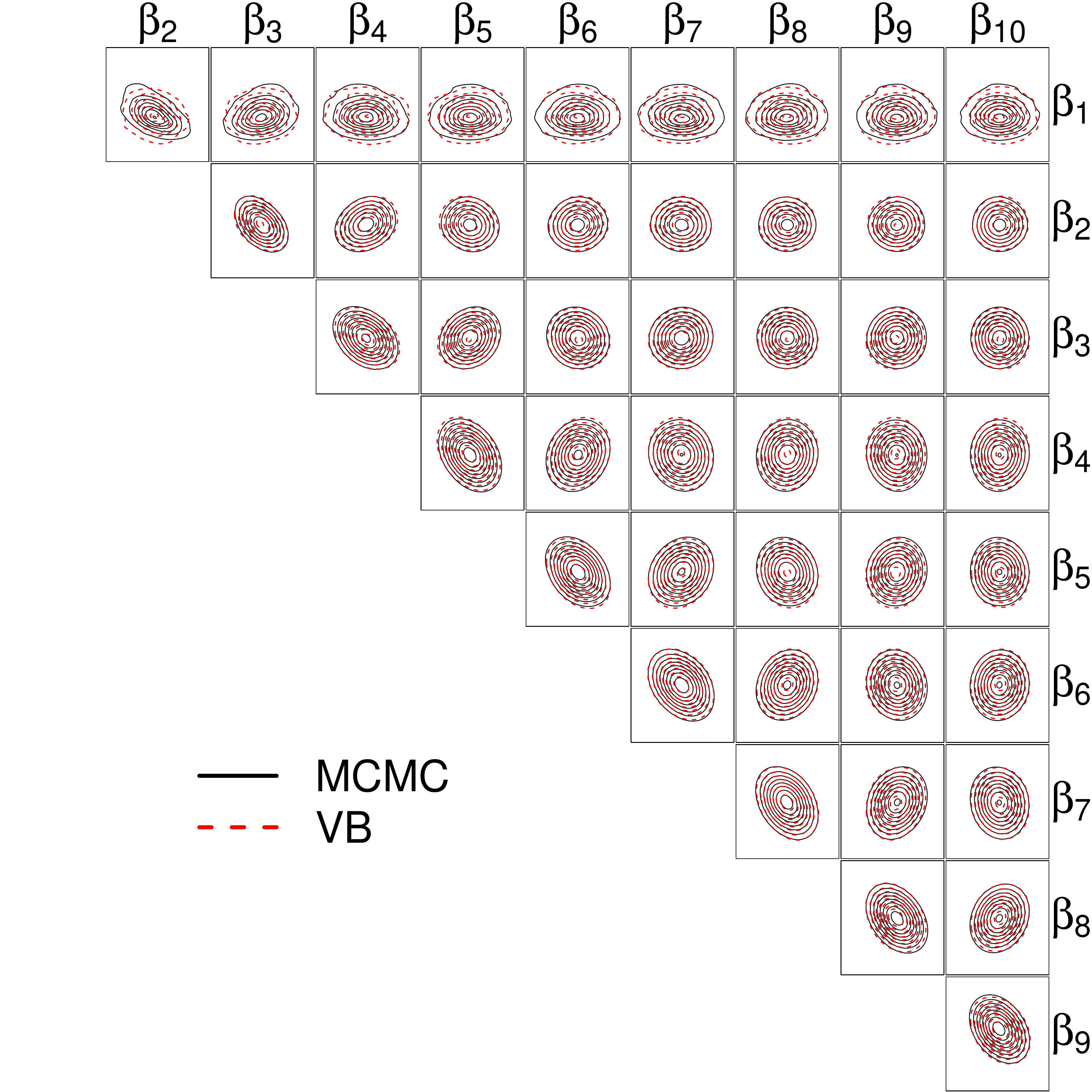}
    \caption{ADVI and MCMC bivariate joint posteriors for the first 10 B-spline coefficients.}
    \label{fig:joint_betas}
\end{figure}

\begin{figure}[H]
    \centering
    \includegraphics[scale=0.22]{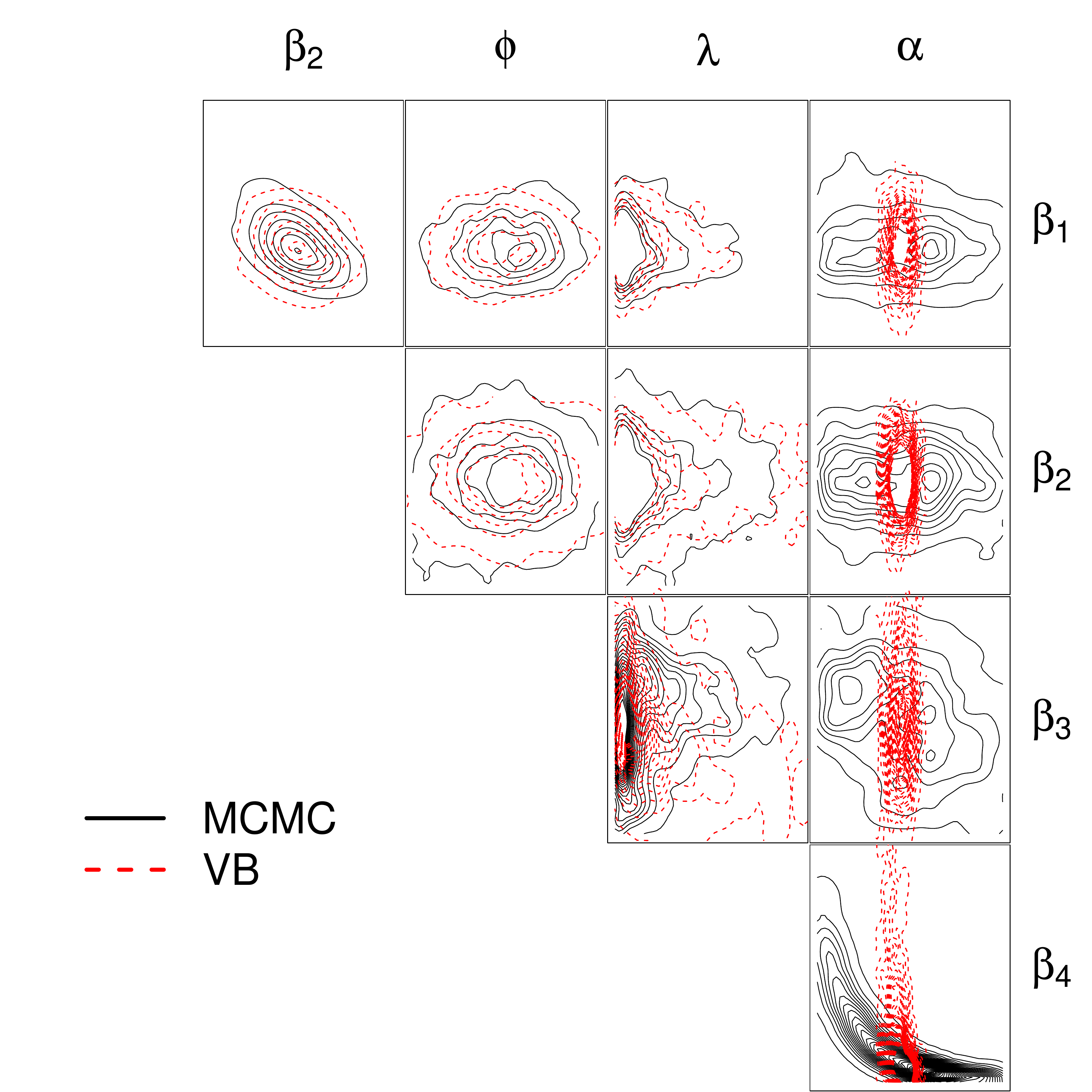}
    \caption{ADVI and MCMC bivariate joint posteriors for $\beta_1, \beta_2, \phi, \lambda, \alpha$.}
    \label{fig:joint_betas_phi_lambda_alpha}
\end{figure}

Figure \ref{fig:marginals_vb_mcmc} shows that both methods produce very similar posterior marginal distributions for the spline coefficients and that the marginals capture the true values of the coefficients. Figure \ref{fig:lambda_alpha_phi_vb_mcmc} shows MCMC and ADVI posterior marginals for $\lambda$, $\phi$ and $\alpha$. The marginal posterior distributions for $\phi$ and $\lambda$ under VB and MCMC are similar, while $\alpha$ posterior estimates via VB are less dispersed in comparison with MCMC marginal posterior densities. Finally, joint posterior uncertainties are also very similar under both aproaches, as examplified in Figures \ref{fig:joint_betas} and \ref{fig:joint_betas_phi_lambda_alpha}, except for the bridge penalization parameter $\alpha$. Despite the discrepancies regarding estimation of $\alpha$ by the VB and MCMC methods, there is very little difference regarding the goodness of fit to the simulated data (\ref{fig:conf_bands_first_replica}) which might indicate that the data brings little information about $\alpha$.

Furthermore, similar conclusions can be drawn from the other 99 replicates. Posterior point estimates produced by the proposed ADVI and by the baseline MCMC on all 100 replicas are compared in Figure \ref{fig:point_estimates_all_replicas} and Figure \ref{fig:point_estimates_all_replicas_APP} in the appendix. Point estimates for $\beta_k, \ k=1, \ldots, 34$ are very similar under both methods, except for $\beta_{34}.$ However, a closer look reveals that the 34-th column of the B-spline regression matrix $\bfX$ has all entries equal to 0, except the last one, which makes $\beta_{34}$ very hard to estimate as it has very little impact on the mean function $f_{\beta}(x).$ The level of agreement between the point estimates produced by MCMC and the proposed ADVI for $\lambda$ and $\alpha$ are not as high as for the regression coefficients $\beta_k, \ k=1, \ldots, 34$.

\begin{figure*}
    \centering
    \includegraphics[scale=0.8]{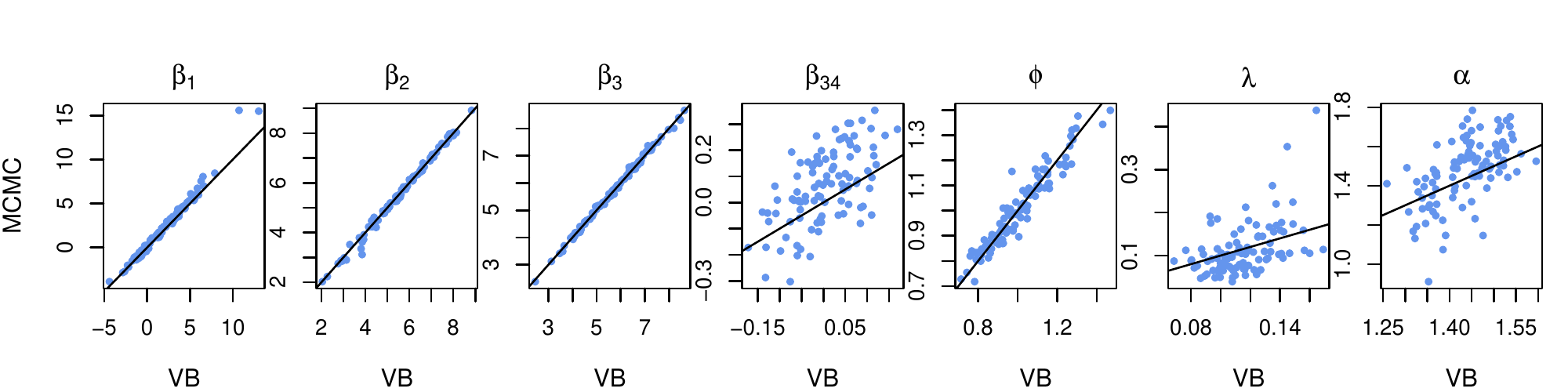}
    \caption{Point estimates for all parameters under MCMC and VB for each one of the 100 replicas.  The black lines represent the identity function. The scatterplots for the remaining coefficients can be seen in Figure \ref{fig:point_estimates_all_replicas_APP} in the appendix.}
    \label{fig:point_estimates_all_replicas}
\end{figure*}

\subsection{Scenario 2: varying sample sizes}

\begin{table*}
\begin{center}
\begin{minipage}{\textwidth}
  \caption{Number of batches ($n_b$), number of epochs ($n_e$) and total number of iterations ($n_i$) of the proposed VB algorithm according to the sample sizes ($n$) of the simulated datasets.}\label{tab:specifications_mcmc_vb} 
\centering
\begin{tabular}{@{\extracolsep{5pt}} ccccccc} 
\\[-1.8ex]\hline 
\hline \\[-1.8ex] 
$n$  & 1,000 & 10,000 & 50,000 & 100,000 & 500,000 & 1,000,000 \\ \hline
$n_{b}$ & 1,000 & 1,000 & 1,000 & 10,000 & 10,000 & 10,000 \\ 
$n_{e}$ & 2,000 & 200 & 100 & 100 & 100 & 100 \\ 
$n_{i}$ & 2,000 & 2,000 & 5,000 & 1,000 & 5,000 & 10,000 \\ 
\botrule
\end{tabular} 
\end{minipage}
\end{center}
\end{table*} 

In this section, the effect of sample sizes in computational times of the MCMC and ADVI algorithms in the context of estimation of the semi-parametric Bayesian bridge is investigated, focusing on large datasets. The specifications for the simulated data are the same as described in section \ref{sec:simulation_1}, except that now the sample sizes of the simulated data vary as $n\in \{10^3, \ 10^4, \ 5\times \ 10^4, \ 10^5, \ 5\times \ 10^5, \ 10^6\}.$ For all six cases, the MCMC was ran for 5000 iterations (except for n=1000, which required 50000 iterations until we could get reasonable evidence of convergence), initializing $\bfbeta$ and $\phi$ at their OLS estimates under no regularization. For the VB implementation, we always used learning rate 0.01 for Adam and $M=100$ Monte Carlo samples to approximate expectations in the calculation of the ELBO. Distinct batch sizes and number of epochs were for each sample size according to Table \ref{tab:specifications_mcmc_vb}, which implies 1000 to 10000 iterations in total depending on the value of $n$ to be compared with the results obtained after 5000 iterations of MCMC.

Regarding computational time, the proposed ADVI algorithm for semi-parametric Bayesian bridge regression is orders of magnitude faster than MCMC, as shown in Table \ref{tab:computational_times}. ADVI remais faster even when running more iterations than MCMC and also when VB times are adjusted to account for the 5000 iterations ran under MCMC.

In summary, the proposed VB aproach is capacble of reaching convergence faster than MCMC, using less epochs and providing an accurate approximation of the posterior. 

\begin{table}[!htbp] 
\centering 
  \caption{Computational times (in seconds) for MCMC and ADVI implementations according to the size of the simulated data ($n$). *: In this case we ran 50000 mcmc iterations instead of 5000 because of lack of evidence for convergence under MCMC.} 
  \label{tab:computational_times} 
\begin{tabular}{@{\extracolsep{5pt}} cccc} 
\\[-1.8ex]\hline 
\hline \\[-1.8ex] 
$n$ & MCMC & ADVI & ADVI \\ 
 &  &  & \small{(5000 iterations)}\\ 
\hline \\[-1.8ex] 
$10^3$ & $219s^*$ & $6s$ & $15s$\\ 
$10^4$ & $64s$ & $6s$ & $15s$ \\ 
$5\times 10^4$ & $294s$ & $15s$ & $15s$\\ 
$10^5$ & $618s$ & $13s$ & $65s$\\ 
$5\times 10^5$ & $2,744s$ & $76s$ & $76s$ \\ 
$10^6$ & $4,570s$ & $143s$ & $72s$\\ 
\hline \\[-1.8ex] 
\end{tabular} 
\end{table} 

\subsection{Scenario 3: multiple covariates with non-parametric effects}
\label{sec:scenario3}

In this scenario, a dataset with $n=1000$ observations $(y_i, x_{i1}, x_{i2}), \ i = 1, \ldots, n$ were simulated from the model $y_i = \beta_0 + f_1(x_{i1}; \tau_1) + f_2(x_{i2}; \tau_2) + \epsilon_i, \ \epsilon_i\sim N(0, \sigma^2),$ where $f_i( \cdot ; \tau_i)$ denotes a realization of a 1-dimensional Gaussian process with mean 0 and covariance function $Cov: \mathbb{R}\times \mathbb{R}\rightarrow\mathbb{R}, \  Cov(z, z') = \exp\left[-\tau^{-1}_i (z - z')^2 \right]$. In the simulation, we used $\tau_1 = 1$ and $\tau_2 = 2$. The objective is to investigate how the proposed semi-parametric bridge regression model performs when there are more than one covariate affecting the response in a non-parametric way. The results presented here assume $n_1 = n_2 = 100$ knots for each b-spline non-parametric effect.

\begin{figure*}
    \includegraphics[scale=0.35]{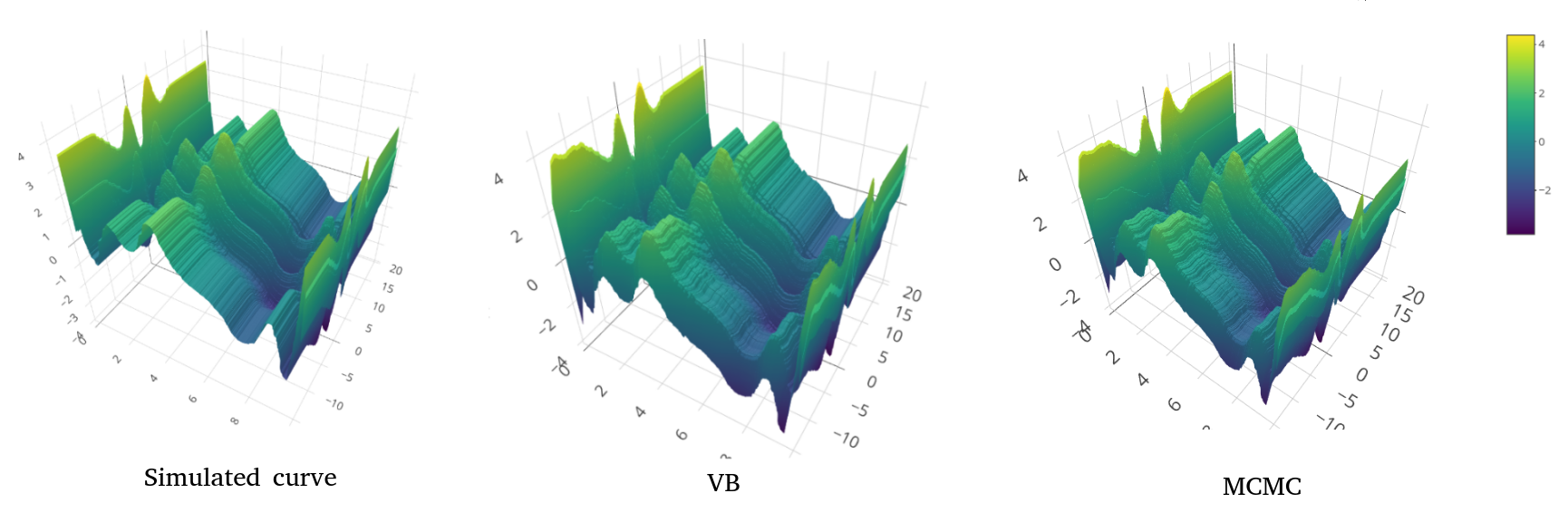}
    \caption{Simulated mean of $y$ as a function of covariates $x_1$ and $x_2$ according to section \ref{sec:scenario3}. Posterior point estimate for the average of $y$ under the proposed MCMC approach.}
    \label{fig:persp}
\end{figure*}

Figure \ref{fig:persp} shows that the underlying simulated surface for the mean of the response variable $y$ as a function of $x_1$ and $x_2$ is accurately estimated by the MCMC and VB approaches.  It is worth to mention that the adopted VB is much faster than MCMC. 

\begin{figure*}
    \centering
    \includegraphics[scale=0.7]{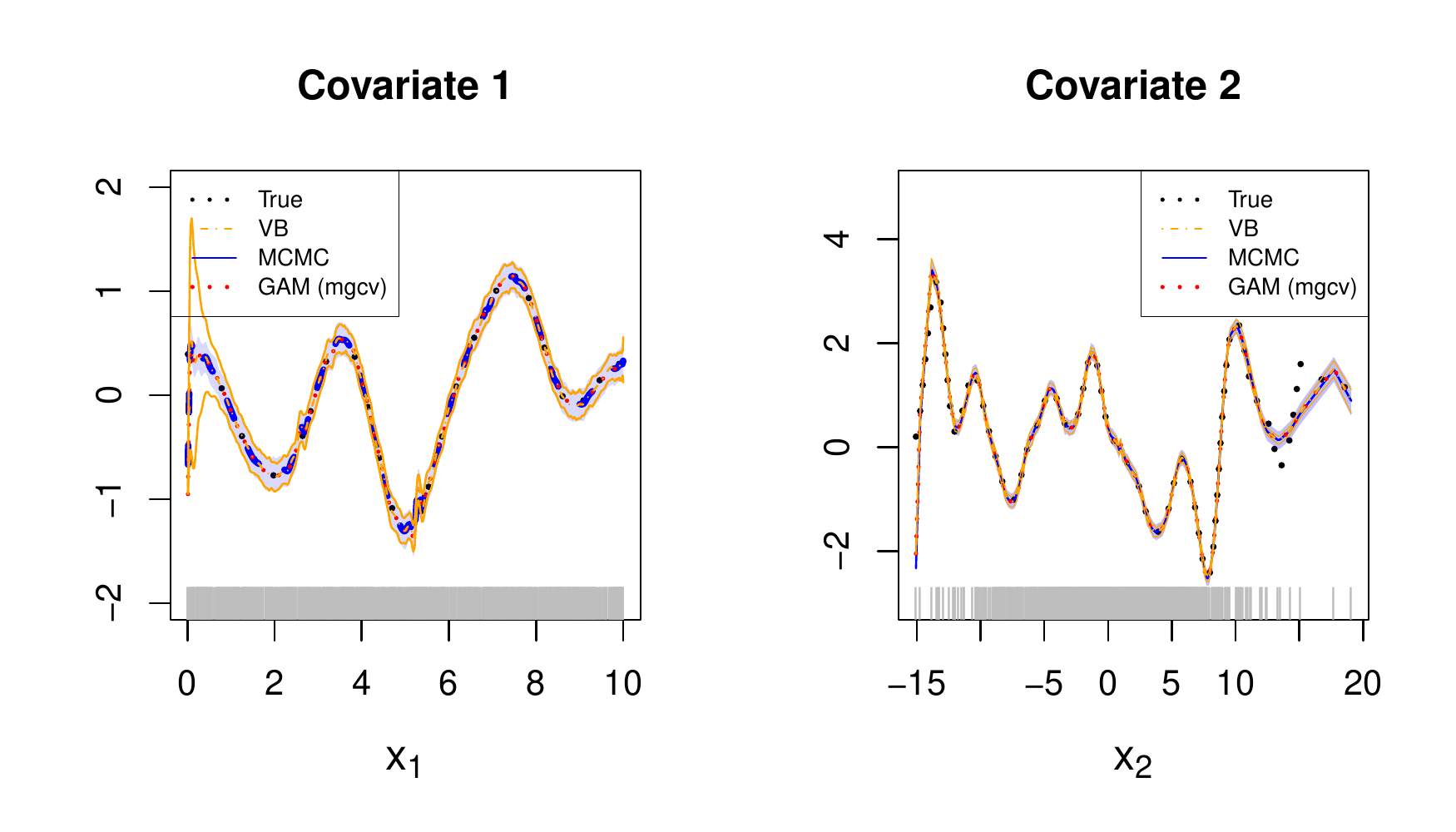}
    \caption{Comparison of MCMC and GAM when fitting the simulated data from section \ref{sec:scenario3}.}
    \label{fig:gam_vc_mcmc_multi_covariates}
\end{figure*}

When comparing the model formulation under MCMC, VB, and the point estimates from GAM, Figure \ref{fig:gam_vc_mcmc_multi_covariates} shows that they yield  virtually indistinguishable estimates for the non-parametric effects. While GAM penalizes the second-order differences of the basis functions' coefficients, the bridge penalization used in the MCMC and VB leads to the same fit.

\section{Real data application}
\label{sec:real_data}

\begin{figure*}
    \centering
    \includegraphics[scale=0.45]{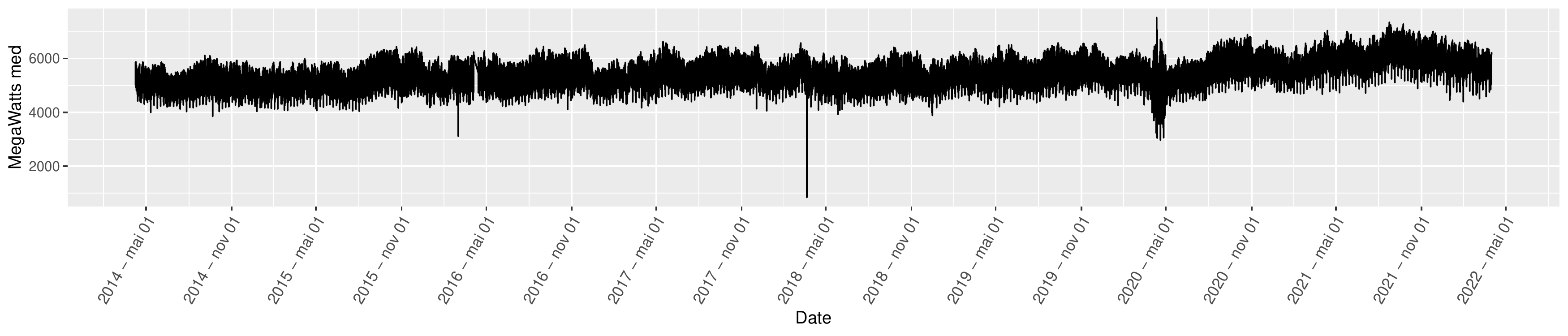}
    \caption{Full Energy Charge data (measured hourly).}
    \label{fig:dataset_energy_full}
\end{figure*}

This section illustrates the use of the proposed ADVI inference procedure for semi-parametric Bayesian bridge regression on a large real dataset. The data consists of hourly measured Energy Charges starting from 2014-04-06 (10pm) to 2022-03-31 (11pm) averaged over stations in the Northern region of Brazil (see Figure \ref{fig:dataset_energy_full}). In total, there are $n=69717$ obsevations. The dataset is maintained by Operador Nacional do Sistema Elétrico (ONS) and can be obtained at  https://dados.ons.org.br/dataset/carga-energia .

The data exhibits strong seasonal patterns with multiple frequencies due to periodicities in energy consumption according with time of the day, day of the week, season of the year and possibly more. We model the seasonal harmonics as the cosine and sine Fourier basis representation for weekly periodicity (period = $24\times 7=168$) following \cite{west2006}. To capture overall level changes in the series, we included a cubic B-spline with one knot at every 100 hours for a total of 700 knots. The B-spline coefficients are subjected to bridge penalization while the coefficients of the Fourier harmonics basis functions are not penalized. In total, the resulting covariate matrix has $p=868$ columns.

\begin{figure*}
    \centering
    \includegraphics[scale=0.45]{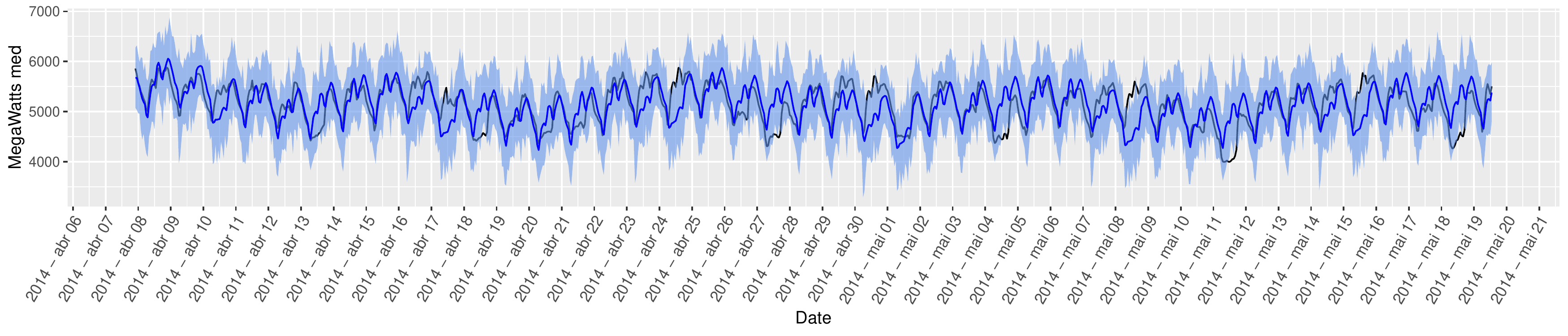}
    \caption{VB 95\% credibility bands and posterior mean for the average response. Only the first 1000 observations of Energy Charge data are shown.}
    \label{fig:dataset_energy_first}
\end{figure*}

\begin{figure*}
    \centering
    \includegraphics[scale=0.45]{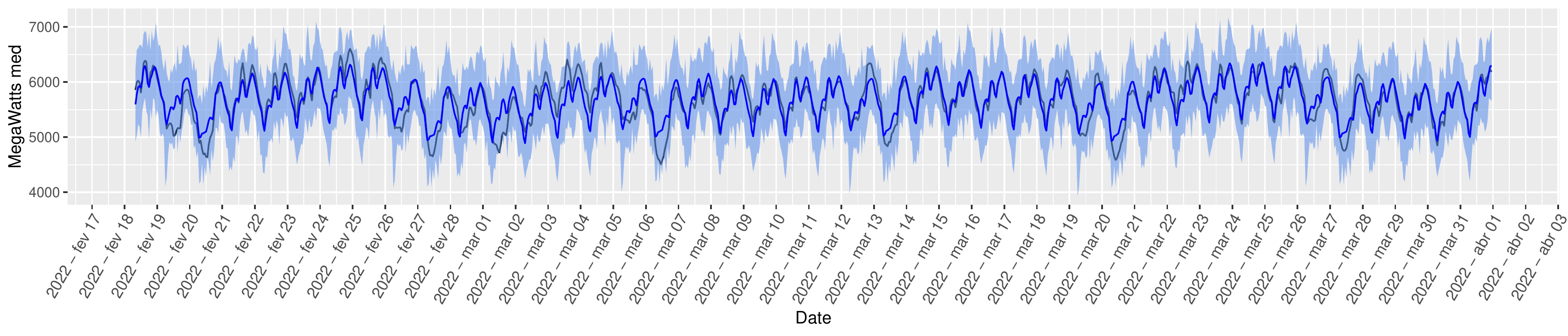}
    \caption{VB 95\% credibility bands and posterior mean for the average response. Only the last 1000 observations of Energy Charge data are shown.}
    \label{fig:dataset_energy_last}
\end{figure*}

The proposed ADVI inference approach took approximately 831 seconds (less than 14min) to complete 2000 iterations using 100 Monte Carlo samples to estimate the gradient of the ELBO. On the other hand, MCMC takes 69239 seconds (19.23 hours) to run the same 2000 iterations. Posterior estimates for the beginning and end of the series are shown in Figures \ref{fig:dataset_energy_first} and \ref{fig:dataset_energy_last}.

\section{Conclusions and future work }
\label{sec:conclusion}

The present work developed a variational inference procedure based on ADVI for Bayesian inference in semi-parametric bridge regression models. The use of small batches of data at each iteration of the training algorithm   reduces computational time in comparison with a more traditional MCMC approach. Full Bayesian inference is preserved so joint uncertainty estimates for all model parameters are available. It was verified in the simulation study that the joint posterior is well approximated by the proposed variational family. 

Some directions for future research include (i) the extension to non-Gaussian distribtutions for the response variable; (ii) expanding flexibility of the variational family with other approaches such as the semi implicit variational inference from \cite{yin2018} and (iii) considering different spline basis functions to represent non-parametric effects of covariaties including GAM models with tensor splines avoiding MCMC and RJMCMC. See details in \cite{denison1998}, \cite{dias2002}, \cite{Li2013} and  references therein.

\section{Acknowledgements}
This paper was partially supported by Fapesp Grants (R. Dias) 2018/04654-9, (R. Dias and H. S. Migon) 2019/10800-0 and (R. Dias) 2019/00787-7.

\begin{appendices}

\section{Details for the MCMC algorithm}
\label{sec:MCMC_details}

\subsection{Full conditionals}
\label{sec:gibbs}

The original posterior distribution (without variable augmentation) is given by 

\begin{multline}
p(\bfbeta, \bfbeta_0, \bflambda, \phi, \bfalpha \mid \bfy) \propto p(\bfy \mid \bfbeta, \bfbeta_0, \phi)p(\bfbeta_0)\times \\ 
\times p(\bfbeta \mid \bflambda, \phi, \bfalpha)p(\bflambda)p(\phi)p(\bfalpha),
\label{eq:gibbs_joint_unmarginalized}
\end{multline}

\noindent where $\bflambda = (\lambda_1, \ldots, \lambda_D)$ and $\bfalpha = (\alpha_1, \ldots, \alpha_D)$.

The joint posterior distribution with the uniform-gamma variable augmentation is expressed as

\begin{multline*}
p(\bfbeta, \bfbeta_0, \bflambda, \phi, \bfu, \bfalpha \mid \bfy) \propto p(\bfy \mid \bfbeta, \bfbeta_0, \phi)p(\bfbeta_0)\times\\
\times p(\bfbeta \mid \bfu, \phi, \bfalpha)p(\bfu\mid \bflambda, \bfalpha)p(\bflambda)p(\phi)p(\bfalpha).
\end{multline*}

\noindent With $\bfalpha$ fixed, all full conditional distributions are analytically available, therefore a straightforward Gibbs sampler scheme can be implemented. The procedure from \cite{Damien_Walker_2012} is used to get a numerically stable sampler for the truncated Gaussian and truncated exponential distributions that appear in the Gibbs sampler scheme that follows. \\

\noindent \textbf{Posterior full conditional distribution of $\bfbeta_j$}
%\begin{multline*}
%    p(\bfbeta \mid \bfy, \bfgamma, \phi, \bfu, \alpha) = \\
%    N\left(\bfbeta; \ (\XtX)^{-1}\bfX^{\top}(\bfy-\bfZ\bfgamma), \right. \\ \left. (\XtX)^{-1}\phi^{-1} \right)\prod^{p_x}_{j=1} \mathds{1}\left(\mid\beta_j\mid < u^{\frac{1}{\alpha}}_j\phi^{-\frac{1}{2}} \right),
%\end{multline*}

%\noindent where $N(\bfx; \bfmu, \bfSigma)\mathds{1}(\bfx \in \mathcal{A})$ denotes the multivariate Gaussian distribution with mean $\bfmu$ and covariance matrix $\bfSigma$, truncated in the set $\mathcal{A}$ and evaluated at $\bfx$.\\

\begin{multline*}
    (\bfbeta_j \mid \bfy, \bfbeta_0, \phi, \bfu, \bfalpha) \sim \\
    N\left(\ ( \bfX^{\top}_j \bfX^{\top}_j)^{-1}\bfX^{\top}_j(\bfy-\bfX_0\bfbeta_0 - \sum_{k \neq j}\bfX_k \bfbeta_k), \right.\\ 
    \left. (\bfX^{\top}_j \bfX_j)^{-1}\phi^{-1} \right) \mathds{1}_{\mathcal{B}_j},
\end{multline*}

\noindent where \begin{multline*}\mathcal{B}_j = \left\{ (\beta_{j1}, \ldots, \beta_{j,K_j}) \in \mathbb{R}^{K_j}: \right. \\ \left.\lvert \beta_{jk}\rvert < u^{\frac{1}{\alpha_j}}_{jk}\phi^{-\frac{1}{2}}, \ \forall k \in \{1, \ldots, K_j\} \right\}\end{multline*} and $N(\bfmu, \bfSigma)\mathds{1}_{\mathcal{A}}$ denotes the multivariate Gaussian distribution with mean $\bfmu$ and covariance matrix $\bfSigma$, truncated in the set $\mathcal{A}$.\\

\noindent \textbf{Posterior full conditional distribution of $\bfbeta_0$}

\begin{multline*}
    (\bfbeta_0\mid \bfy, \phi, \bfbeta) \sim N\left( \left( \bfX^{\top}_0\bfX_0\phi + \Sigma^{-1}_{0}\right)^{-1} \times \right. \\ \left. \times \left( \phi \bfX^{\top}_0\left(\bfy - \sum^D_{j = 1}\bfX_j\bfbeta_j\right) + \Sigma^{-1}_{0}\bfmu_{0} \right) \right., \\ \left.\left( \bfX^{\top}_0\bfX_0\phi + \Sigma^{-1}_{0}\right)^{-1} \right)
\end{multline*}

\noindent \textbf{Posterior full conditional distribution of $\phi$}

\begin{multline*}
    (\phi\mid \bfy, \bfbeta, \bfbeta_0, \bfu, \bfalpha) \sim \\ \sim Ga \left( \frac{n}{2}+ \frac{1}{2}\sum^D_{j=1}K_j + a_{\phi}, \ \frac{1}{2}RSS(\bfbeta, \bfbeta_0)+b_{\phi} \right) \\ \mathds{1}_{\left( 0, \ \min_{j,  \ell }\left( u^{ \frac{2}{\alpha_j} }_{j\ell} \mid\beta_{j\ell}\mid^{-2}\right)\right)}
\end{multline*}

\noindent where $RSS(\bfbeta, \bfbeta_0)=(\bfy - \sum_{j=1}^D\bfX_j\bfbeta_j - \bfX_0\bfbeta_0)^{\top}(\bfy - \sum_{j=1}^D\bfX_j\bfbeta_j - \bfX_0\bfbeta_0)$ denotes the residual sum of squares and $Ga(a, b)\mathds{1}_{\mathcal{A}}$ denotes the Gamma distribution with mean $a/b$ and variance $a/b^2$ truncated in the set $\mathcal{A}$.\\

\noindent \textbf{Posterior full conditional distribution of $u_{j\ell}$}

\begin{equation*}
    (u_{j\ell} \mid \bfy, \lambda_j, \bfbeta, \phi, \alpha_j) \sim Exp(\lambda_j)\mathds{1}_{ \left(\mid\beta_{j\ell}\mid^{\alpha_j}\phi^{\frac{\alpha_j}{2}}, \ +\infty \right) },
\end{equation*}
\noindent where $Exp(\lambda)\mathds{1}_{\mathcal{A}}$ denotes the exponential distribution with rate parameter $\lambda$ truncated on $\mathcal{A}$.\\

\noindent \textbf{Posterior conditional distribution of $\lambda$}\\

When sampling from $\lambda_j$, we can marginalize out $\bfu_j$. Such marginalization was found to improve mixing of the Markov chains. 

\begin{multline*}
    (\lambda_j \mid \bfy, \bfbeta_j, \phi, \alpha_j) \sim \\ \sim Ga\left(a_{\lambda} + \frac{K_j}{\alpha_j}, \ b_{\lambda} + \phi^{\frac{\alpha_j}{2}}\sum^{K_j}_{\ell=1} \mid \beta_{j\ell} \mid^{\alpha_j}\right).
\end{multline*}

\subsection{Estimation of $\alpha_j$}

There is no possible choice of prior for $\alpha_j$ that leads to an analytically available full conditional distribution on $\alpha_j$. Therefore a Metropolis-Hastings (MH) step is proposed for $\alpha_j$. The MH algorithm requires specification of a proposal distribution $q(\alpha^*_j \vert \alpha^{(i)}_j)$ where $\alpha^{(i)}_j$ denotes the value of $\alpha_j$ in the current iteration $i$ and $\alpha^*_j$ denotes the proposed value for $\alpha_j$ according to the proposal density $q(\cdot \mid \alpha^{(i)}_j)$.

\subsubsection{Marginalized MH proposal}
\label{sec:MH_alpha}

When sampling $\alpha_j$, we consider the reduced parameter vector  $\bftheta=(\bfalpha, \bfbeta, \bfbeta_0, \phi, \bflambda)$ marginalizing out the auxiliary variables $\bfu$. Alternatively, appendix \ref{append:MH} describes a proposal $q(\cdot \mid \alpha^{(i)}_j)$ based on the full parameter vector $(\bfalpha, \bfbeta, \bfbeta_0, \phi, \bflambda, \bfu)$ including the augmented variables $\bfu$. It was found crucial to marginalize $\bfu$ out of the model when sampling $\alpha_j$ in the MCMC. 

We denote by $\bftheta^*=(\alpha^*_j, \bfalpha_{-j}, \bfbeta, \bfbeta_0, \phi, \bflambda)$ the non-augmented parameter vector at the proposed $\alpha_j = \alpha^*_j$, with $\bfalpha_{-j}$ denoting the vector $\alpha$ excluding its $j$-th entry and by $\bftheta=(\bfalpha, \bfbeta, \bfbeta_0, \phi, \bflambda)$ the parameter vector with $\alpha_j$ representing the current iteration. Notice that  the auxiliary variables $\bfu$ are marginalized out. The proposed $\alpha^*_j$ is accepted with probability $\rho(\alpha^*_j \mid \alpha_j) = \min\{1, r(\alpha^*_j \mid \alpha_j)\}$ where 
\begin{align*}
r(\alpha^*_j \mid \alpha_j) &= \frac{p(\bfy \mid \bftheta^*)p(\bftheta^*)q(\alpha_j \mid \alpha^*_j)}{p(\bfy \mid \bftheta)p(\bftheta)q(\alpha^{*}_j \mid \alpha_j)} \\
&= \frac{p(\bfbeta_j \mid \lambda_j, \phi, \alpha^*_j)p(\alpha^*_j)q(\alpha_j \mid \alpha^*_j)}{p(\bfbeta_j \mid \lambda_j, \phi, \alpha_j)p(\alpha_j)q(\alpha^{*}_j \mid \alpha_j)}.
\end{align*}

We chose a prior on $\alpha_j$ given by $\alpha_j = 2.5\eta_j$ where $\eta_j\sim Beta(a_{\eta}, b_{\eta} )$. The MH proposal $q(\alpha^{*}_j \mid \alpha_j)$ is derived from a Gaussian random walk on $v = \log \frac{\alpha_j}{2.5-\alpha_j}$, i.e., $q(v^* \mid v) = N(v^*; v, \ W)$ where $N(x; \ \mu, \sigma^2)$ denotes the Gaussian density with mean $\mu$ and variance $\sigma^2$ evaluated at $x$. It is easy to show that the Gaussian random walk $q(v^* \mid v)$ on $v$ implies $q(\alpha^*_j \mid \alpha_j) = N \left( \log\frac{\alpha^*_j}{2.5 - \alpha^*_j}; \ \log\frac{\alpha_j}{2.5 - \alpha_j}, W_j\right)\times \frac{2.5}{\alpha^*_j(2.5-\alpha^*_j)}.$\\

After simplifications, 
\begin{align*}
r&(\alpha^*_j \mid \alpha_j) = \frac{{\alpha^*_j}^{K_j}}{\alpha^{K_j}_j} \times \frac{\Gamma(\alpha^{-1}_j )^{K_j}}{\Gamma({\alpha^*_j}^{-1} )^{K_j}} \times \lambda_j^{\frac{K_j}{\alpha^*_j} - \frac{K_j}{\alpha_j} } \times\\
&\times \exp\left\{ -\lambda_j \sum^{K_j}_{\ell=1} \left(\mid\beta_{j\ell}\mid^{\alpha^*_j} \phi^{\frac{\alpha^*_j}{2}} - \mid\beta_{j\ell}\mid^{\alpha} \phi^{\frac{ \alpha }{2}} \right) \right\} \times \\
&\times \frac{ {\alpha^*_j}^{a_{\eta} } (2.5-\alpha^*_j)^{b_{\eta}} }{\alpha^{a_{\eta}}_j (2.5-\alpha_j)^{b_{\eta} }}.
\end{align*}

\subsubsection{Non-marginalized MH proposal}
\label{append:MH}

Section \ref{sec:MH_alpha} described a MH scheme for $\alpha_j$ taking the advantage of the marginalization of the auxiliary variables $\bfu.$ This section describes a simpler alternative that does not marginalize $\bfu$. However, we could not get well mixing Markov chains by usig the simpler proposal described here.

We chose a prior on $\alpha_j$ given by $\alpha_j = 2.5\eta_j$ where $\eta_j\sim Beta(a_{\eta}, b_{\eta} )$. The MH proposal $q(\alpha^{*}_j \vert \alpha_j)$ is derived from a Gaussian random walk on $v = \log \frac{\alpha_j}{2.5-\alpha_j}$, i.e., $q(v^* \mid v) = N(v^*; v, W)$ where $N(x; \mu, \sigma^2)$ denotes the Gaussian density with mean $\mu$ and variance $\sigma^2$ evaluated at $x$. It is easy to show that the Gaussian random walk $q(v^* \mid v)$ on $v$ implies $q(\alpha^*_j \mid \alpha_j) = N \left( \log\frac{\alpha^*_j}{2.5 - \alpha^*_j}; \ \log\frac{\alpha_j}{2.5 - \alpha_j}, W\right)\times \frac{2.5}{\alpha^*_j(2.5-\alpha^*_j)}.$

After simplifications,
\begin{align*}
r(\alpha^*_j \mid \alpha_j) &= \lambda_j^{\left( \frac{K_j}{\alpha^*_j} - \frac{K_j}{\alpha_j} \right)}\times \frac{\Gamma\left(\frac{1}{\alpha^*_j} + 1\right)^{K_j}}{ \Gamma\left( \frac{1}{\alpha_j} + 1 \right)^{K_j}}\times \\
&\times\frac{ {\alpha^*_j}^{a_{\eta} } (2.5-\alpha^*_j)^{b_{\eta}} }{  {\alpha_j}^{a_{\eta}} (2.5-\alpha_j)^{b_{\eta} }}\mathds{1}(m \leq \alpha^*_j \leq M),
\end{align*}
where $$m=\max\left( \{0\} \cup  \left\{ \frac{\log(u_{jk}) }{\log (\mid\beta_{jk}\mid  \phi^{1/2} )}: \ \ k \in S_{-} \right\} \right),$$ $$M=\min\left( \{2.5\} \cup \left\{ \frac{\log(u_{jk}) }{\log( \mid\beta_{jk}\mid \phi^{1/2} )}: \ \ k \in S_{+} \right\} \right),$$ $$S_{-}: = \{j=1, \ldots, K_j: \ 0 < \mid\beta_{jk}\mid< \phi^{-1/2}\},$$ $$S_{+}: = \{j=1, \ldots, K_j: \ \mid\beta_{jk}\mid > \phi^{-1/2} \}.$$

It is important to notice that $S_{-}$ does not include values of $k$ such that $\beta_{jk}=0$.

\section{Details of BBVI for semi-parametric Bayesian bridge}
\label{append:bbvi}

The details for implementation of BBVI for the proposed semi-parametric Bayesian bridge are described in this section. Only the case with $D=1$ is shown, although the calculations could be easily extended to the multivariate case where $D\in \mathbb{N}$.

Under mean-field, the proposed marginal variational distributions are $q(\bfbeta \mid m'_{\beta}, S_{\beta}) = N(\bfbeta \mid m'_{\beta}, S_{\beta} ), \
q(\gamma\mid m'_{\gamma}, S_{\gamma}) = N(\bfgamma \mid m'_{\gamma}, S_{\gamma} ),     \
q(\phi\mid a'_{\phi}, b'_{\phi}) = Ga(\phi \mid a'_{\phi}, b'_{\phi}), \
q(\lambda \mid a'_{\lambda}, b'_{\lambda}) = Ga(\lambda \mid a'_{\lambda}, b'_{\lambda}), \
q(\alpha\mid a'_{\alpha}, b'_{\alpha}) = 2\times Beta(\alpha \mid a'_{\alpha}, b'_{\alpha}).$ 

We now describe the analytical expressions for the gradient of $\log q_{\bfpsi}(\bftheta)$ with respect to each variational parameter. In the equations bellow, $\mbox{dig}(\cdot)$ denotes the digamma function, i.e., the derivative of the log gamma function.

\begin{align*}
\nabla_{S^{-1}_{\beta}} \log q(\bfbeta \mid m'_{\beta}, S_{\beta}) &=- \frac{1}{2}(\bfbeta - m'_{\beta})(\bfbeta - m'_{\beta})^{\top}\\
&{\color{white}{.}} \ \ \ \ \ \ \ \ \ \ \ \ \ \ \ \ \ \ \ \ \ \ \ \ \ \ \ \ + \frac{1}{2}S_{\beta}\\ 
\nabla_{m'_{\beta}} \log q(\bfbeta \mid m'_{\beta}, S_{\beta}) &= S^{-1}_{\beta}(\bfbeta - m'_{\beta})\\
\nabla_{S^{-1}_{\gamma}} \log q(\bfgamma \mid m'_{\gamma}, S_{\gamma}) &= \frac{1}{2}S_{\gamma} - \frac{1}{2}(\bfgamma - m'_{\gamma})(\bfgamma - m'_{\gamma})^{\top}\\
\nabla_{m'_{\gamma}} \log q(\bfgamma \mid m'_{\gamma}, S_{\gamma}) &= S^{-1}_{\gamma}(\bfgamma - m'_{\gamma})\\
\nabla_{a'_{\phi}} \log q(\phi \mid a'_{\phi}, b'_{\phi}) &= \log b'_{\phi} - \mbox{dig}(a'_{\phi}) + \log \phi\\
\nabla_{b'_{\phi}} \log q(\phi \mid a'_{\phi}, b'_{\phi}) &= \frac{a'_{\phi}}{b'_{\phi}} - \phi\\
\nabla_{a'_{\lambda}} \log q(\lambda \mid a'_{\lambda}, b'_{\lambda}) &= \log b'_{\lambda} - \mbox{dig}(a'_{\lambda}) + \log \lambda\\
\nabla_{b'_{\lambda}} \log q(\lambda \mid a'_{\lambda}, b'_{\lambda}) &= \frac{a'_{\lambda}}{b'_{\lambda}} - \lambda\\
\nabla_{a'_{\alpha}} \log q(\alpha \mid a'_{\alpha}, b'_{\alpha}) &= -\log 2 + \mbox{dig}(a'_{\alpha} + b'_{\alpha}) \\
&- \mbox{dig}(a'_{\alpha}) + \log \alpha\\
\nabla_{b'_{\alpha}} \log q(\alpha \mid a'_{\alpha}, b'_{\alpha}) &= -\log 2 + \mbox{dig}(a'_{\alpha} + b'_{\alpha}) \\ 
&- \mbox{dig}(b'_{\alpha}) + \log (2 - \alpha)
\end{align*}

\section{Variational family in Bayes bridge reparameterization method}

The details for implementation of the reparameterization method for the proposed semi-parametric Bayesian bridge are described in this section. Only the case with $D=1$ is shown, although the calculations could be easily extended to the multivariate case where $D\in \mathbb{N}$.

For the Bayesian lasso model, we have $\bftheta = (\bfbeta^{\top}, \bfgamma^{\top}, \lambda, \phi)^{\top}$ as the parameter vector. The proposed marginal variational distributions are $q(\bfbeta \mid m'_{\beta}, S_{\beta}) = N(\bfbeta \mid m'_{\beta}, S_{\beta} ), \
q(\gamma\mid m'_{\gamma}, S_{\gamma}) = N(\bfgamma \mid m'_{\gamma}, S_{\gamma} ),     \
q(\phi\mid a'_{\phi}, b'_{\phi}) = \log N(\phi \mid a'_{\phi}, b'_{\phi}), \
q(\lambda \mid a'_{\lambda}, b'_{\lambda}) = \log N(\lambda \mid a'_{\lambda}, b'_{\lambda}).$ The proposed joint variational distribution follows the mean-field assumption, i.e. $q(\bftheta) = q(\bfbeta)q(\bfgamma)q(\lambda)q(\phi).$

The entropy of multivariate normal distributions and log-normal distributions are available in closed form. If $q(\bftheta) = N_d(\bftheta; \ \bfm, \bfS),$ where $N_d(\bftheta; \ \bfm, \bfS)$ denotes the density of a $d$-dimensional multivariate normal distribution with mean vector $\bfm$ and covariance matrix $\bfS$ evaluated at $\bftheta,$ the entropy of $q(\bftheta)$ is $H(q(\bftheta); \bfm, \bfS) = \frac{1}{2}\log \mid2\pi e \bfS\mid = \frac{d}{2}[ \log(2\pi) + 1] + \sum^d_{i=1} \ell_{ii},$ where $\ell_{ii}$ denotes the $i$-th entry of the diagonal of $\bfL,$ the Choleskey decomposition of $\bfS.$ For the log-Normal distribution, if $q(\theta) = \log N(\theta; m, s),$ meaning $\log \theta \sim N(m, s^2),$ then $H( q(\theta); m, s) = \log_2\left(s\sqrt{2\pi}e^{\mu+0.5}\right).$

\section{Detailed expression for gradient of the ELBO under ADVI}
\label{sec:ADVI_details}

This section derives in details the approximation of the gradient of the ELBO in equation \eqref{eq:elbo_advi_stochastic_gradient} to facilitate implementation of Algorithm 1.

Let $\epsilon^{(\ell)} \sim N(\bf0_d, \bfI_d),$ $\ell = 1, \ldots, M$ where $d = K_0 + K_1 + \ldots + K_D + 2D + 1$ is the dimension of the parameter vector $\bftheta$. Then we compute $\bfxi^{(\ell)} = \bfm + \bfL \bfepsilon^{(\ell)}$ and using the transformation $T$ defined in section \ref{sec:advi}, we compute the implied Monte Carlo samples in the original parameter vector $\bftheta$ as \begin{align*}
\bfbeta_0^{(\ell)} &= \bfxi_{\bfbeta_0^{(\ell)}}, \\
\bfbeta_d^{(\ell)} &= \bfxi_{\bfbeta_d^{(\ell)}}, \ d = 1, \ldots, D\\
\phi^{(\ell)} &= e^{\xi_{{\phi}^{(\ell)}}}, \\
\lambda_d^{(\ell)} &= e^{\xi_{\lambda_d^{(\ell)}}}, \ d = 1, \ldots, D\\
\alpha_d^{(\ell)} &= \frac{2.5}{1 + e^{-\xi_{{\alpha_d^{(\ell)}}}} }, \ d = 1, \ldots, D.
\end{align*}

Starting from equation \eqref{eq:elbo_advi_stochastic_gradient},  we get

\begin{multline}
\widetilde{\nabla}_{\bfpsi} ELBO(\tilde{\bfy},\bfpsi) = \\
= \frac{n}{KM} \sum^M_{\ell=1} \left[ \nabla_{\bfpsi} \log p(\tilde{\bfy} \mid T^{-1}(\bfxi^{(\ell)}) ) \right] \ + \ \\ + \frac{1}{M} \sum^M_{\ell=1} \nabla_{\bfpsi}\left[ \log p(T^{-1}(\bfxi^{(\ell)}) ) + \log \mid J_{T^{-1}(\bfxi^{(\ell)})} \mid \right] \\
+ \nabla_{\bfpsi}\mathbb{E}_{\bfxi \sim N(\bfm, \bfL\bfL^{T})} \left[- \log N(\bfxi ; \ \bfm, \bfL\bfL^{\top})\right]\\
= \\
\frac{n}{KM}\sum^M_{\ell = 1} \nabla_{\bfpsi} \log N\left(\widetilde{\bfy}; \ \sum^D_{j=0}\widetilde{X}_j \bfbeta^{(\ell)}_j, {\phi^{(\ell)}}^{-1}\bfI \right)+\\
\frac{1}{M}\sum^M_{\ell = 1}\nabla_{\bfpsi}\left[ \log p(\bfbeta^{(\ell)}_0) + \log p(\phi^{(\ell)}) + \right.\\
\left. + \sum^D_{j = 1} \sum^{K_j}_{k = 1} \log p(\bfbeta^{(\ell)}_{jk} \vert \lambda^{(\ell)}_j, \phi^{(\ell)}, \alpha^{(\ell)}_j) + \right.\\
+ \left. \sum^D_{j=1}\log p(\lambda^{(\ell)}_j) + \sum^D_{j=1} \log p(\alpha^{(\ell)}_j)\right] + \\
+\frac{1}{M} \sum^M_{\ell = 1} \nabla_{\bfpsi} \bfxi_{\phi}^{(\ell)} +\\
+\frac{1}{M} \sum^M_{\ell = 1} \sum^D_{j=1} \nabla_{\bfpsi}\left[  \xi^{(\ell)}_{\lambda_j} + \xi^{(\ell)}_{\alpha_j} - 2 \log\left(1 + e^{\xi^{(\ell)}_{\alpha_j}}\right) \right]+\\
+ \nabla_{\bfpsi} \sum^{d}_{i = 1}\bfL_{ii}\\
= \\
\frac{n}{KM}\sum^M_{\ell = 1} \nabla_{\bfpsi} \log N\left(\widetilde{\bfy}; \  \sum^D_{j=0}\widetilde{X}_j \bfbeta^{(\ell)}_j, {\phi^{(\ell)}}^{-1}\bfI \right)+\\
\frac{1}{M}\sum^M_{\ell = 1}\nabla_{\bfpsi} \log N(\bfbeta^{(\ell)}_0; \bfmu_0, \bfSigma_0) + \\ 
\frac{1}{M}\sum^M_{\ell = 1}\nabla_{\bfpsi}\log Ga(\phi^{(\ell)}; a_{\phi}, b_{\phi}) + \\
 + \frac{1}{M}\sum^M_{\ell = 1}\sum^D_{j = 1} \sum^{K_j}_{k = 1} \nabla_{\bfpsi} \log GG\left( \bfbeta^{(\ell)}_{jk};  0, \frac{ {\lambda_j^{(\ell)}}^{-\frac{1}{\alpha^{(\ell)}_j}}}{\sqrt{\phi^{(\ell)}} }, \alpha^{(\ell)}_j \right) + \\
+ \frac{1}{M}\sum^M_{\ell = 1}\sum^D_{j=1}\nabla_{\bfpsi}\log Ga(\lambda^{(\ell)}_j; a_{\lambda}, b_{\lambda}) + \\
\frac{1}{M}\sum^M_{\ell = 1}\sum^D_{j=1} \nabla_{\bfpsi}\log {\alpha^{(\ell)}_j}^{a_{\eta} - 1} (2.5 - \alpha^{(\ell)}_j)^{b_\eta - 1} + \\
+\frac{1}{M} \sum^M_{\ell = 1} \nabla_{\bfpsi}\log\phi^{(\ell)} + \frac{1}{M} \sum^M_{\ell = 1} \sum^D_{j=1} \nabla_{\bfpsi} \log\lambda^{(\ell)}_j + \\
+\frac{1}{M} \sum^M_{\ell = 1} \sum^D_{j=1}\nabla_{\bfpsi}\left(\log \frac{\alpha^{(\ell)}_j}{2.5 - \alpha^{(\ell)}_j} - 2 \log \frac{2.5}{2.5 - \alpha^{(\ell)}_j} \right)+\\
+ \nabla_{\bfpsi} \sum^{d}_{i = 1}\bfL_{ii},
\end{multline}

\noindent where the term $\nabla_{\bfpsi} \sum^{d}_{i = 1}\bfL_{ii}$ comes from the fact that the entropy of a multivariate Gaussian random vector $\bfx \sim N(\bfmu, \bfSigma)$ is $\mathbb{H}[\bfx] = \mathbb{E}_{\bfx \sim N(\bfmu, \bfSigma)}[-\log N(\bfx \mid \bfmu, \bfSigma)] = \log\vert 2\pi e \bfSigma\vert$.

\section{Further results and comparisons}

This section presents more results on the simulation described in section \ref{sec:simulation_1}. 

\begin{figure*}
    \centering
    \includegraphics[scale=0.75]{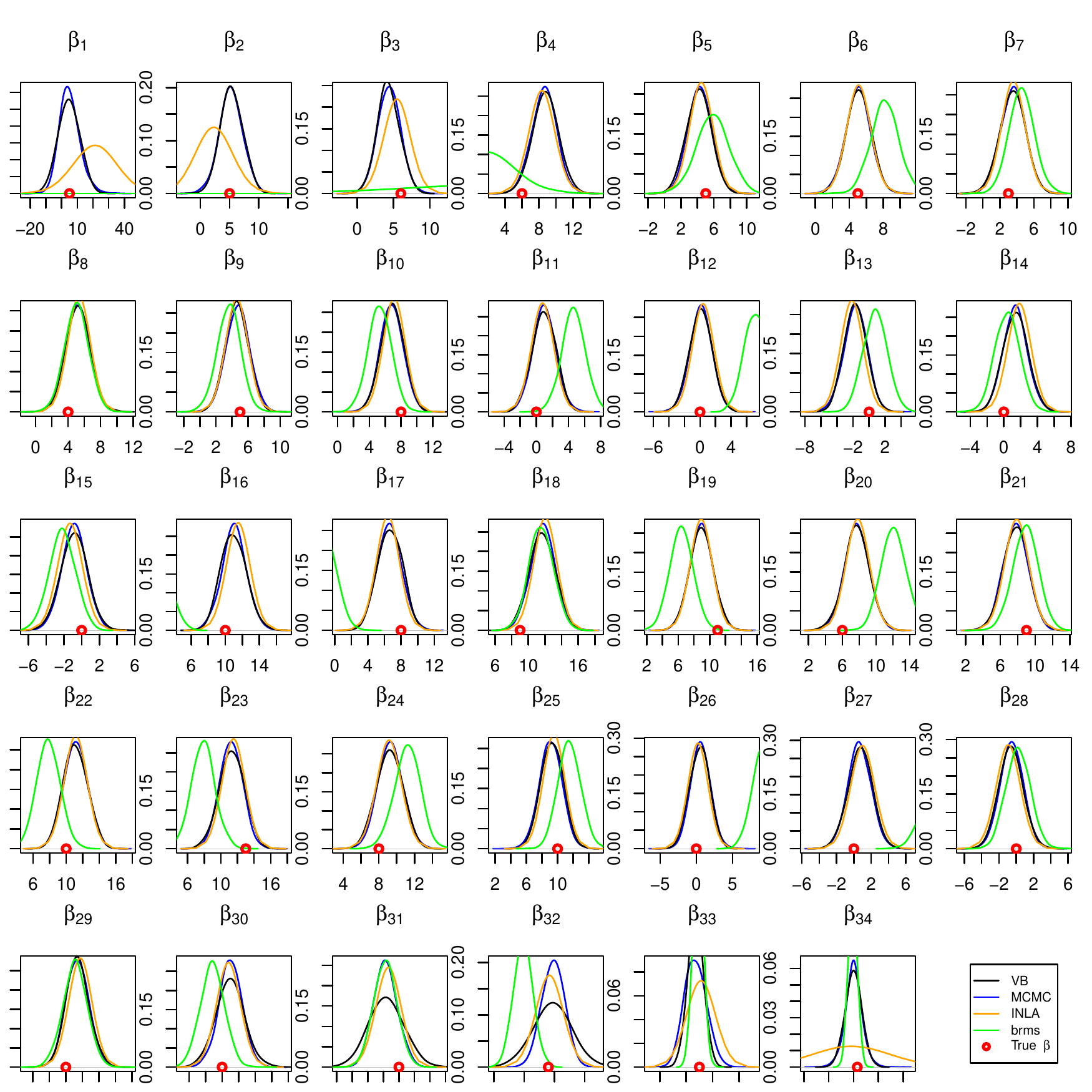}
    \caption{Marginal posterior distributions for $\beta_k, \ k=1, \ \ldots, 34$ based on the first replicate of the simulated data. Red dots denote the true values of the parameters.}
    \label{fig:marginals_vb_mcmc_APP}
\end{figure*}

Figure \ref{fig:marginals_vb_mcmc_APP} contains the posterior marginal distributions for all parameters of the model described in section \ref{sec:simulation_1}. The marginals for $\beta_1, \ldots, \beta_7$ are also shown in Figure \ref{fig:marginals_vb_mcmc}. As in Figure \ref{fig:marginals_vb_mcmc}, Figure \ref{fig:marginals_vb_mcmc_APP} shows high level of agreement between marginals obtained via MCMC and ADVI, with brms having higher discrepancy with respect to the MCMC. INLA also approximates well the marginal distributions except for a small number of coefficients (e.g., $\beta_1, \beta_2, \beta_{34}$)

\begin{figure*}
    \centering
    \includegraphics[scale=0.85]{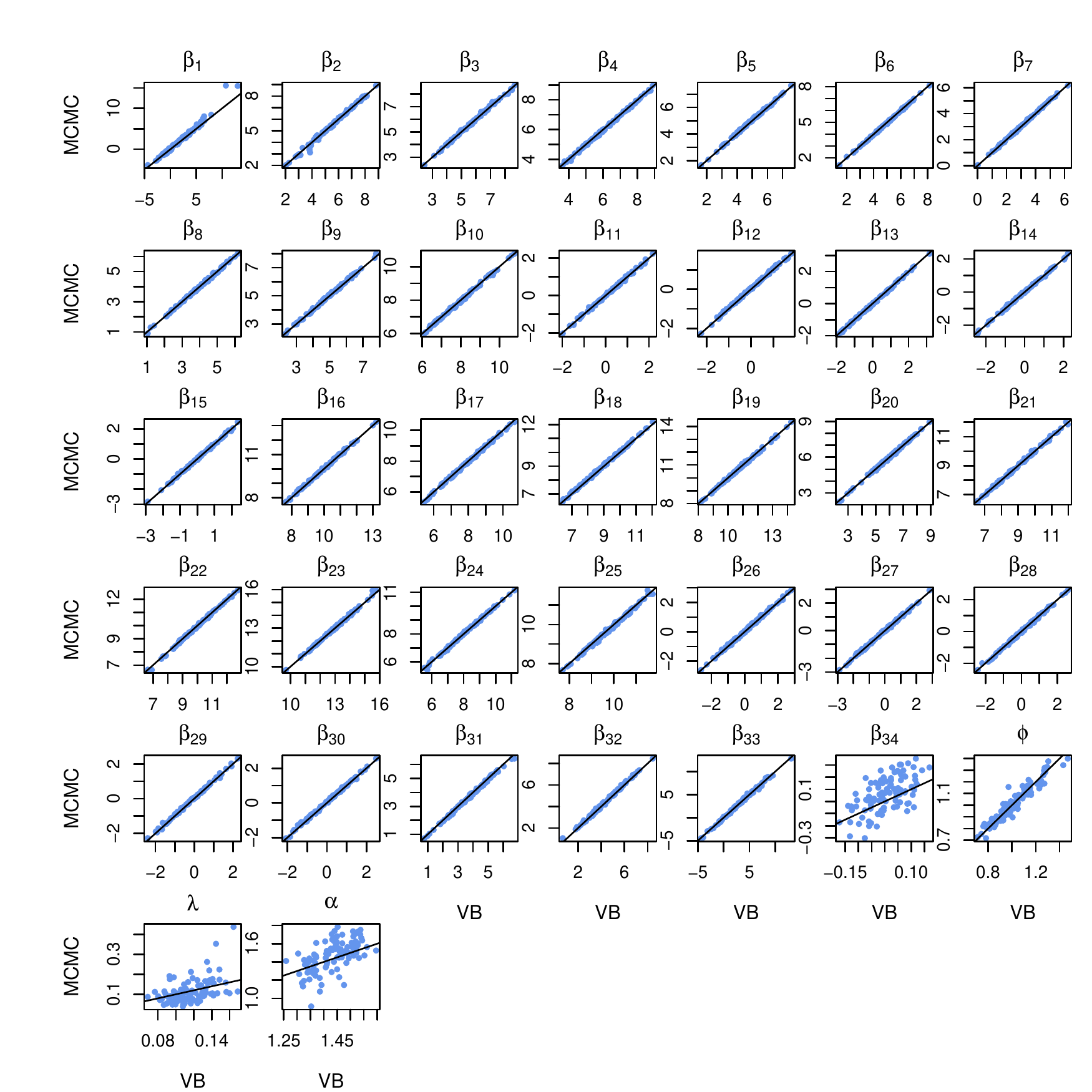}
    \caption{Point estimates for all parameters under MCMC and VB for each one of the 100 replicas.  The black lines represent the identity function.}
    \label{fig:point_estimates_all_replicas_APP}
\end{figure*}

Figure \ref{fig:point_estimates_all_replicas_APP} complements Figure  \ref{fig:point_estimates_all_replicas} from section \ref{sec:simulation_1} with the point estimates for all parameters obtained via ADVI and MCMC for each one of the 100 simulated datasets.

\begin{figure*}
    \centering
    \includegraphics[scale=0.5]{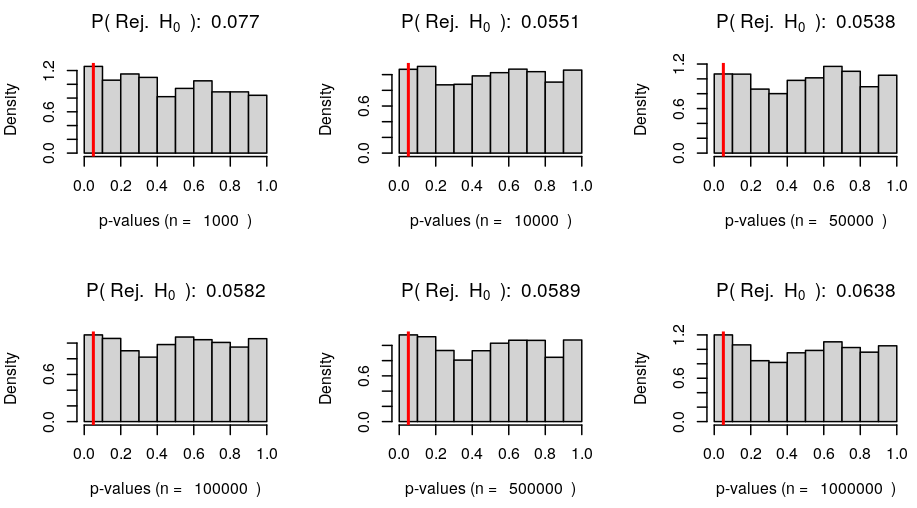}
    \caption{p-values for the Kolmogorov-Smirnov test for comparison of each marginal posterior predictive distributions under ADVI and MCMC. In red, we highlight P(type 1 error) = 5\%. The expression P(Rej. $H_0$) represents the proportion of times we would reject the null hypothesis of equal distributions under ADVI and MCMC if we were to reject whenever p-value $<$ 0.05.}
    \label{fig:hist_pvalues}
\end{figure*}

To further address the quality of the variational approximations to the true posterior distribution, Figure \ref{fig:hist_pvalues} shows the p-values obtained when comparing (via Kolmogorov-Smirnov tests) marginal posterior predictive distributions estimated by VB and MCMC. The shapes of the histograms are approximately uniform with a low prevalence of small p-values, as expected under the null hypothesis that the distributions under MCMC and VB are the same. For example, the empirical proportions of p-values below 0.05 are close to 0.05, which corresponds to the expected proportion of false discoveries when $H_0$ is true and a type 1 error of 0.05 is fixed.

\newpage

\end{appendices}

\newpage

\bibliography{vb_bridge.bib}% common bib file

\begin{thebibliography}{}
\providecommand{\doi}[1]{\url{https://doi.org/#1}}
\bibcommenthead

\bibitem [\protect \citeauthoryear {%
Alves%
, Dias%
\BCBL {}\ \BBA {} Migon%
}{%
Alves%
\ \protect \BOthers {.}}{%
{\protect \APACyear {2021}}%
}]{%
alves2021}
\APACinsertmetastar {%
alves2021}%
\begin{APACrefauthors}%
Alves, L.%
, Dias, R.%
\BCBL {} Migon, H.S.%
\end{APACrefauthors}%
\unskip\
\newblock
\APACrefYearMonthDay{2021}{}{}.
\newblock
{\BBOQ}\APACrefatitle {Variational Full Bayes Lasso: Knots Selection in
  Regression Splines} {Variational full bayes lasso: Knots selection in
  regression splines}.{\BBCQ}
\newblock
\APACjournalVolNumPages{arXiv preprint arXiv:2102.13548}{}{}{}.
\newblock

\newblock

\PrintBackRefs{\CurrentBib}

\bibitem [\protect \citeauthoryear {%
Armagan%
}{%
Armagan%
}{%
{\protect \APACyear {2009}}%
}]{%
armagan2009}
\APACinsertmetastar {%
armagan2009}%
\begin{APACrefauthors}%
Armagan, A.%
\end{APACrefauthors}%
\unskip\
\newblock
\APACrefYearMonthDay{2009}{}{}.
\newblock
{\BBOQ}\APACrefatitle {Variational bridge regression} {Variational bridge
  regression}.{\BBCQ}
\newblock
 \APACrefbtitle {Artificial Intelligence and Statistics} {Artificial
  intelligence and statistics}\ (\BPGS\ 17--24).
\PrintBackRefs{\CurrentBib}

\bibitem [\protect \citeauthoryear {%
Blei%
, Kucukelbir%
\BCBL {}\ \BBA {} McAuliffe%
}{%
Blei%
\ \protect \BOthers {.}}{%
{\protect \APACyear {2017}}%
}]{%
blei2017}
\APACinsertmetastar {%
blei2017}%
\begin{APACrefauthors}%
Blei, D.M.%
, Kucukelbir, A.%
\BCBL {} McAuliffe, J.D.%
\end{APACrefauthors}%
\unskip\
\newblock
\APACrefYearMonthDay{2017}{}{}.
\newblock
{\BBOQ}\APACrefatitle {Variational inference: A review for statisticians}
  {Variational inference: A review for statisticians}.{\BBCQ}
\newblock
\APACjournalVolNumPages{Journal of the American statistical
  Association}{112}{518}{859--877}.
\newblock

\newblock

\PrintBackRefs{\CurrentBib}

\bibitem [\protect \citeauthoryear {%
Bürkner%
}{%
Bürkner%
}{%
{\protect \APACyear {2017}}%
}]{%
brms}
\APACinsertmetastar {%
brms}%
\begin{APACrefauthors}%
Bürkner, P\BHBI C.%
\end{APACrefauthors}%
\unskip\
\newblock
\APACrefYearMonthDay{2017}{}{}.
\newblock
{\BBOQ}\APACrefatitle {{brms}: An {R} Package for {Bayesian} Multilevel Models
  Using {Stan}} {{brms}: An {R} package for {Bayesian} multilevel models using
  {Stan}}.{\BBCQ}
\newblock
\APACjournalVolNumPages{Journal of Statistical Software}{80}{1}{1--28}.
\newblock

\newblock

\PrintBackRefs{\CurrentBib}

\bibitem [\protect \citeauthoryear {%
Carvalho%
, Polson%
\BCBL {}\ \BBA {} Scott%
}{%
Carvalho%
\ \protect \BOthers {.}}{%
{\protect \APACyear {2009}}%
}]{%
carvalho2009}
\APACinsertmetastar {%
carvalho2009}%
\begin{APACrefauthors}%
Carvalho, C.M.%
, Polson, N.G.%
\BCBL {} Scott, J.G.%
\end{APACrefauthors}%
\unskip\
\newblock
\APACrefYearMonthDay{2009}{}{}.
\newblock
{\BBOQ}\APACrefatitle {Handling sparsity via the horseshoe} {Handling sparsity
  via the horseshoe}.{\BBCQ}
\newblock
 \APACrefbtitle {Artificial Intelligence and Statistics} {Artificial
  intelligence and statistics}\ (\BPGS\ 73--80).
\PrintBackRefs{\CurrentBib}

\bibitem [\protect \citeauthoryear {%
Casella%
, Ghosh%
, Gill%
\BCBL {}\ \BBA {} Kyung%
}{%
Casella%
\ \protect \BOthers {.}}{%
{\protect \APACyear {2010}}%
}]{%
Casella_2010}
\APACinsertmetastar {%
Casella_2010}%
\begin{APACrefauthors}%
Casella, G.%
, Ghosh, M.%
, Gill, J.%
\BCBL {} Kyung, M.%
\end{APACrefauthors}%
\unskip\
\newblock
\APACrefYearMonthDay{2010}{}{}.
\newblock
{\BBOQ}\APACrefatitle {{Penalized regression, standard errors, and Bayesian
  lassos}} {{Penalized regression, standard errors, and Bayesian
  lassos}}.{\BBCQ}
\newblock
\APACjournalVolNumPages{Bayesian Analysis}{5}{2}{369 -- 411}.
\newblock
\begin{APACrefURL} {https://doi.org/10.1214/10-BA607} \end{APACrefURL}
\newblock

\newblock

\newblock
\begin{APACrefDOI} \doi{10.1214/10-BA607} \end{APACrefDOI}
\PrintBackRefs{\CurrentBib}

\bibitem [\protect \citeauthoryear {%
Currie%
\ \BBA {} Durban%
}{%
Currie%
\ \BBA {} Durban%
}{%
{\protect \APACyear {2002}}%
}]{%
currie2002}
\APACinsertmetastar {%
currie2002}%
\begin{APACrefauthors}%
Currie, I.D.%
\BCBT {}\ \BBA {} Durban, M.%
\end{APACrefauthors}%
\unskip\
\newblock
\APACrefYearMonthDay{2002}{}{}.
\newblock
{\BBOQ}\APACrefatitle {Flexible smoothing with P-splines: a unified approach}
  {Flexible smoothing with p-splines: a unified approach}.{\BBCQ}
\newblock
\APACjournalVolNumPages{Statistical Modelling}{2}{4}{333--349}.
\newblock

\newblock

\PrintBackRefs{\CurrentBib}

\bibitem [\protect \citeauthoryear {%
Damien%
\ \BBA {} Walker%
}{%
Damien%
\ \BBA {} Walker%
}{%
{\protect \APACyear {2001}}%
}]{%
Damien_Walker_2012}
\APACinsertmetastar {%
Damien_Walker_2012}%
\begin{APACrefauthors}%
Damien, P.%
\BCBT {}\ \BBA {} Walker, S.G.%
\end{APACrefauthors}%
\unskip\
\newblock
\APACrefYearMonthDay{2001}{}{}.
\newblock
{\BBOQ}\APACrefatitle {Sampling Truncated Normal, Beta, and Gamma Densities}
  {Sampling truncated normal, beta, and gamma densities}.{\BBCQ}
\newblock
\APACjournalVolNumPages{Journal of Computational and Graphical
  Statistics}{10}{2}{206-215}.
\newblock

\newblock

\PrintBackRefs{\CurrentBib}

\bibitem [\protect \citeauthoryear {%
Denison%
, Mallick%
\BCBL {}\ \BBA {} Smith%
}{%
Denison%
\ \protect \BOthers {.}}{%
{\protect \APACyear {1998}}%
}]{%
denison1998}
\APACinsertmetastar {%
denison1998}%
\begin{APACrefauthors}%
Denison, D.%
, Mallick, B.%
\BCBL {} Smith, A.%
\end{APACrefauthors}%
\unskip\
\newblock
\APACrefYearMonthDay{1998}{}{}.
\newblock
{\BBOQ}\APACrefatitle {Automatic {B}ayesian curve fitting} {Automatic
  {B}ayesian curve fitting}.{\BBCQ}
\newblock
\APACjournalVolNumPages{Journal of the Royal Statistical Society: Series B
  (Statistical Methodology)}{60}{2}{333--350}.
\newblock

\newblock

\PrintBackRefs{\CurrentBib}

\bibitem [\protect \citeauthoryear {%
Dias%
\ \BBA {} Gamerman%
}{%
Dias%
\ \BBA {} Gamerman%
}{%
{\protect \APACyear {2002}}%
}]{%
dias2002}
\APACinsertmetastar {%
dias2002}%
\begin{APACrefauthors}%
Dias, R.%
\BCBT {}\ \BBA {} Gamerman, D.%
\end{APACrefauthors}%
\unskip\
\newblock
\APACrefYearMonthDay{2002}{}{}.
\newblock
{\BBOQ}\APACrefatitle {A {B}ayesian approach to hybrid splines non-parametric
  regression} {A {B}ayesian approach to hybrid splines non-parametric
  regression}.{\BBCQ}
\newblock
\APACjournalVolNumPages{Journal of Statistical Computation and
  Simulation}{72}{4}{285--297}.
\newblock

\newblock

\PrintBackRefs{\CurrentBib}

\bibitem [\protect \citeauthoryear {%
Duchi%
, Hazan%
\BCBL {}\ \BBA {} Singer%
}{%
Duchi%
\ \protect \BOthers {.}}{%
{\protect \APACyear {2011}}%
}]{%
duchi2011}
\APACinsertmetastar {%
duchi2011}%
\begin{APACrefauthors}%
Duchi, J.%
, Hazan, E.%
\BCBL {} Singer, Y.%
\end{APACrefauthors}%
\unskip\
\newblock
\APACrefYearMonthDay{2011}{}{}.
\newblock
{\BBOQ}\APACrefatitle {Adaptive subgradient methods for online learning and
  stochastic optimization.} {Adaptive subgradient methods for online learning
  and stochastic optimization.}{\BBCQ}
\newblock
\APACjournalVolNumPages{Journal of machine learning research}{12}{7}{}.
\newblock

\newblock

\PrintBackRefs{\CurrentBib}

\bibitem [\protect \citeauthoryear {%
Eilers%
\ \BBA {} Marx%
}{%
Eilers%
\ \BBA {} Marx%
}{%
{\protect \APACyear {1996}}%
}]{%
eilers1996}
\APACinsertmetastar {%
eilers1996}%
\begin{APACrefauthors}%
Eilers, P.H.%
\BCBT {}\ \BBA {} Marx, B.D.%
\end{APACrefauthors}%
\unskip\
\newblock
\APACrefYearMonthDay{1996}{}{}.
\newblock
{\BBOQ}\APACrefatitle {Flexible smoothing with B-splines and penalties}
  {Flexible smoothing with b-splines and penalties}.{\BBCQ}
\newblock
\APACjournalVolNumPages{Statistical science}{11}{2}{89--121}.
\newblock

\newblock

\PrintBackRefs{\CurrentBib}

\bibitem [\protect \citeauthoryear {%
Frank%
\ \BBA {} Friedman%
}{%
Frank%
\ \BBA {} Friedman%
}{%
{\protect \APACyear {1993}}%
}]{%
frank1993}
\APACinsertmetastar {%
frank1993}%
\begin{APACrefauthors}%
Frank, L.E.%
\BCBT {}\ \BBA {} Friedman, J.H.%
\end{APACrefauthors}%
\unskip\
\newblock
\APACrefYearMonthDay{1993}{}{}.
\newblock
{\BBOQ}\APACrefatitle {A statistical view of some chemometrics regression
  tools} {A statistical view of some chemometrics regression tools}.{\BBCQ}
\newblock
\APACjournalVolNumPages{Technometrics}{35}{2}{109--135}.
\newblock

\newblock

\PrintBackRefs{\CurrentBib}

\bibitem [\protect \citeauthoryear {%
Hastie%
\ \BBA {} Tibshirani%
}{%
Hastie%
\ \BBA {} Tibshirani%
}{%
{\protect \APACyear {1986}}%
}]{%
hastie1986}
\APACinsertmetastar {%
hastie1986}%
\begin{APACrefauthors}%
Hastie, T.%
\BCBT {}\ \BBA {} Tibshirani, R.%
\end{APACrefauthors}%
\unskip\
\newblock
\APACrefYearMonthDay{1986}{}{}.
\newblock
{\BBOQ}\APACrefatitle {Generalized Additive Models} {Generalized additive
  models}.{\BBCQ}
\newblock
\APACjournalVolNumPages{Statistical Science}{1}{3}{297--310}.
\newblock
\begin{APACrefURL} [{2022-08-19}]{http://www.jstor.org/stable/2245459}
  \end{APACrefURL}
\newblock

\newblock

\PrintBackRefs{\CurrentBib}

\bibitem [\protect \citeauthoryear {%
Hastie%
\ \BBA {} Tibshirani%
}{%
Hastie%
\ \BBA {} Tibshirani%
}{%
{\protect \APACyear {2000}}%
}]{%
hastie2000}
\APACinsertmetastar {%
hastie2000}%
\begin{APACrefauthors}%
Hastie, T.%
\BCBT {}\ \BBA {} Tibshirani, R.%
\end{APACrefauthors}%
\unskip\
\newblock
\APACrefYearMonthDay{2000}{}{}.
\newblock
{\BBOQ}\APACrefatitle {{B}ayesian Backfitting} {{B}ayesian backfitting}.{\BBCQ}
\newblock
\APACjournalVolNumPages{Statistical Science}{15}{3}{196--213}.
\newblock
\begin{APACrefURL} [{2022-06-23}]{http://www.jstor.org/stable/2676659}
  \end{APACrefURL}
\newblock

\newblock

\PrintBackRefs{\CurrentBib}

\bibitem [\protect \citeauthoryear {%
Hastie%
, Tibshirani%
\BCBL {}\ \BBA {} Wainwright%
}{%
Hastie%
\ \protect \BOthers {.}}{%
{\protect \APACyear {2015}}%
}]{%
hastie2015}
\APACinsertmetastar {%
hastie2015}%
\begin{APACrefauthors}%
Hastie, T.%
, Tibshirani, R.%
\BCBL {} Wainwright, M.%
\end{APACrefauthors}%
\unskip\
\newblock
\APACrefYearMonthDay{2015}{}{}.
\newblock
{\BBOQ}\APACrefatitle {Statistical learning with sparsity} {Statistical
  learning with sparsity}.{\BBCQ}
\newblock
\APACjournalVolNumPages{Monographs on statistics and applied
  probability}{143}{}{143}.
\newblock

\newblock

\PrintBackRefs{\CurrentBib}

\bibitem [\protect \citeauthoryear {%
Hoerl%
\ \BBA {} Kennard%
}{%
Hoerl%
\ \BBA {} Kennard%
}{%
{\protect \APACyear {1970}}%
{\protect \APACexlab {{\protect \BCnt {1}}}}}]{%
hoerl1970b}
\APACinsertmetastar {%
hoerl1970b}%
\begin{APACrefauthors}%
Hoerl, A.E.%
\BCBT {}\ \BBA {} Kennard, R.W.%
\end{APACrefauthors}%
\unskip\
\newblock
\APACrefYearMonthDay{1970{\protect \BCnt {1}}}{}{}.
\newblock
{\BBOQ}\APACrefatitle {Ridge regression: applications to nonorthogonal
  problems} {Ridge regression: applications to nonorthogonal problems}.{\BBCQ}
\newblock
\APACjournalVolNumPages{Technometrics}{12}{1}{69--82}.
\newblock

\newblock

\PrintBackRefs{\CurrentBib}

\bibitem [\protect \citeauthoryear {%
Hoerl%
\ \BBA {} Kennard%
}{%
Hoerl%
\ \BBA {} Kennard%
}{%
{\protect \APACyear {1970}}%
{\protect \APACexlab {{\protect \BCnt {2}}}}}]{%
hoerl1970a}
\APACinsertmetastar {%
hoerl1970a}%
\begin{APACrefauthors}%
Hoerl, A.E.%
\BCBT {}\ \BBA {} Kennard, R.W.%
\end{APACrefauthors}%
\unskip\
\newblock
\APACrefYearMonthDay{1970{\protect \BCnt {2}}}{}{}.
\newblock
{\BBOQ}\APACrefatitle {Ridge regression: Biased estimation for nonorthogonal
  problems} {Ridge regression: Biased estimation for nonorthogonal
  problems}.{\BBCQ}
\newblock
\APACjournalVolNumPages{Technometrics}{12}{1}{55--67}.
\newblock

\newblock

\PrintBackRefs{\CurrentBib}

\bibitem [\protect \citeauthoryear {%
Kingma%
\ \BBA {} Ba%
}{%
Kingma%
\ \BBA {} Ba%
}{%
{\protect \APACyear {2014}}%
}]{%
adam}
\APACinsertmetastar {%
adam}%
\begin{APACrefauthors}%
Kingma, D.P.%
\BCBT {}\ \BBA {} Ba, J.%
\end{APACrefauthors}%
\unskip\
\newblock
\APACrefYearMonthDay{2014}{}{}.
\newblock
{\BBOQ}\APACrefatitle {Adam: A method for stochastic optimization} {Adam: A
  method for stochastic optimization}.{\BBCQ}
\newblock
\APACjournalVolNumPages{arXiv preprint arXiv:1412.6980}{}{}{}.
\newblock

\newblock

\PrintBackRefs{\CurrentBib}

\bibitem [\protect \citeauthoryear {%
Kingma%
\ \BBA {} Welling%
}{%
Kingma%
\ \BBA {} Welling%
}{%
{\protect \APACyear {2013}}%
}]{%
kingma2013}
\APACinsertmetastar {%
kingma2013}%
\begin{APACrefauthors}%
Kingma, D.P.%
\BCBT {}\ \BBA {} Welling, M.%
\end{APACrefauthors}%
\unskip\
\newblock
\APACrefYearMonthDay{2013}{}{}.
\newblock
{\BBOQ}\APACrefatitle {Auto-encoding variational {B}ayes} {Auto-encoding
  variational {B}ayes}.{\BBCQ}
\newblock
\APACjournalVolNumPages{arXiv preprint arXiv:1312.6114}{}{}{}.
\newblock

\newblock

\PrintBackRefs{\CurrentBib}

\bibitem [\protect \citeauthoryear {%
Kucukelbir%
, Ranganath%
, Gelman%
\BCBL {}\ \BBA {} Blei%
}{%
Kucukelbir%
\ \protect \BOthers {.}}{%
{\protect \APACyear {2015}}%
}]{%
kucukelbir2015}
\APACinsertmetastar {%
kucukelbir2015}%
\begin{APACrefauthors}%
Kucukelbir, A.%
, Ranganath, R.%
, Gelman, A.%
\BCBL {} Blei, D.%
\end{APACrefauthors}%
\unskip\
\newblock
\APACrefYearMonthDay{2015}{}{}.
\newblock
{\BBOQ}\APACrefatitle {Automatic variational inference in Stan} {Automatic
  variational inference in stan}.{\BBCQ}
\newblock
\APACjournalVolNumPages{Advances in neural information processing
  systems}{28}{}{}.
\newblock

\newblock

\PrintBackRefs{\CurrentBib}

\bibitem [\protect \citeauthoryear {%
Kucukelbir%
, Tran%
, Ranganath%
, Gelman%
\BCBL {}\ \BBA {} Blei%
}{%
Kucukelbir%
\ \protect \BOthers {.}}{%
{\protect \APACyear {2017}}%
}]{%
kucukelbir2017}
\APACinsertmetastar {%
kucukelbir2017}%
\begin{APACrefauthors}%
Kucukelbir, A.%
, Tran, D.%
, Ranganath, R.%
, Gelman, A.%
\BCBL {} Blei, D.M.%
\end{APACrefauthors}%
\unskip\
\newblock
\APACrefYearMonthDay{2017}{}{}.
\newblock
{\BBOQ}\APACrefatitle {Automatic differentiation variational inference}
  {Automatic differentiation variational inference}.{\BBCQ}
\newblock
\APACjournalVolNumPages{The Journal of Machine Learning
  Research}{18}{1}{430--474}.
\newblock

\newblock

\PrintBackRefs{\CurrentBib}

\bibitem [\protect \citeauthoryear {%
Lang%
\ \BBA {} Brezger%
}{%
Lang%
\ \BBA {} Brezger%
}{%
{\protect \APACyear {2004}}%
}]{%
lang2004}
\APACinsertmetastar {%
lang2004}%
\begin{APACrefauthors}%
Lang, S.%
\BCBT {}\ \BBA {} Brezger, A.%
\end{APACrefauthors}%
\unskip\
\newblock
\APACrefYearMonthDay{2004}{}{}.
\newblock
{\BBOQ}\APACrefatitle {{B}ayesian P-splines} {{B}ayesian p-splines}.{\BBCQ}
\newblock
\APACjournalVolNumPages{Journal of computational and graphical
  statistics}{13}{1}{183--212}.
\newblock

\newblock

\PrintBackRefs{\CurrentBib}

\bibitem [\protect \citeauthoryear {%
Leng%
, Tran%
\BCBL {}\ \BBA {} Nott%
}{%
Leng%
\ \protect \BOthers {.}}{%
{\protect \APACyear {2014}}%
}]{%
Leng_2014}
\APACinsertmetastar {%
Leng_2014}%
\begin{APACrefauthors}%
Leng, C.%
, Tran, M\BHBI N.%
\BCBL {} Nott, D.%
\end{APACrefauthors}%
\unskip\
\newblock
\APACrefYearMonthDay{2014}{}{}.
\newblock
{\BBOQ}\APACrefatitle {Bayesian adaptive lasso} {Bayesian adaptive
  lasso}.{\BBCQ}
\newblock
\APACjournalVolNumPages{Annals of the Institute of Statistical
  Mathematics}{66}{}{221--244}.
\newblock

\newblock

\PrintBackRefs{\CurrentBib}

\bibitem [\protect \citeauthoryear {%
F.~Li%
\ \BBA {} Villani%
}{%
F.~Li%
\ \BBA {} Villani%
}{%
{\protect \APACyear {2013}}%
}]{%
Li2013}
\APACinsertmetastar {%
Li2013}%
\begin{APACrefauthors}%
Li, F.%
\BCBT {}\ \BBA {} Villani, M.%
\end{APACrefauthors}%
\unskip\
\newblock
\APACrefYearMonthDay{2013}{}{}.
\newblock
{\BBOQ}\APACrefatitle {Efficient Bayesian Multivariate Surface Regression}
  {Efficient bayesian multivariate surface regression}.{\BBCQ}
\newblock
\APACjournalVolNumPages{Scandinavian Journal of Statistics}{40}{4}{706-723}.
\newblock

\newblock

\PrintBackRefs{\CurrentBib}

\bibitem [\protect \citeauthoryear {%
Q.~Li%
\ \BBA {} Lin%
}{%
Q.~Li%
\ \BBA {} Lin%
}{%
{\protect \APACyear {2010}}%
}]{%
li2010}
\APACinsertmetastar {%
li2010}%
\begin{APACrefauthors}%
Li, Q.%
\BCBT {}\ \BBA {} Lin, N.%
\end{APACrefauthors}%
\unskip\
\newblock
\APACrefYearMonthDay{2010}{}{}.
\newblock
{\BBOQ}\APACrefatitle {The {B}ayesian elastic net} {The {B}ayesian elastic
  net}.{\BBCQ}
\newblock
\APACjournalVolNumPages{{B}ayesian analysis}{5}{1}{151--170}.
\newblock

\newblock

\PrintBackRefs{\CurrentBib}

\bibitem [\protect \citeauthoryear {%
Luts%
\ \BBA {} Wand%
}{%
Luts%
\ \BBA {} Wand%
}{%
{\protect \APACyear {2015}}%
}]{%
luts2015}
\APACinsertmetastar {%
luts2015}%
\begin{APACrefauthors}%
Luts, J.%
\BCBT {}\ \BBA {} Wand, M.P.%
\end{APACrefauthors}%
\unskip\
\newblock
\APACrefYearMonthDay{2015}{}{}.
\newblock
{\BBOQ}\APACrefatitle {Variational inference for count response semiparametric
  regression} {Variational inference for count response semiparametric
  regression}.{\BBCQ}
\newblock

\newblock

\newblock

\PrintBackRefs{\CurrentBib}

\bibitem [\protect \citeauthoryear {%
Mallick%
\ \BBA {} Yi%
}{%
Mallick%
\ \BBA {} Yi%
}{%
{\protect \APACyear {2018}}%
}]{%
mallick2018}
\APACinsertmetastar {%
mallick2018}%
\begin{APACrefauthors}%
Mallick, H.%
\BCBT {}\ \BBA {} Yi, N.%
\end{APACrefauthors}%
\unskip\
\newblock
\APACrefYearMonthDay{2018}{}{}.
\newblock
{\BBOQ}\APACrefatitle {{B}ayesian bridge regression} {{B}ayesian bridge
  regression}.{\BBCQ}
\newblock
\APACjournalVolNumPages{Journal of Applied Statistics}{45}{6}{988--1008}.
\newblock

\newblock

\PrintBackRefs{\CurrentBib}

\bibitem [\protect \citeauthoryear {%
Menictas%
\ \BBA {} Wand%
}{%
Menictas%
\ \BBA {} Wand%
}{%
{\protect \APACyear {2015}}%
}]{%
Menictas2015}
\APACinsertmetastar {%
Menictas2015}%
\begin{APACrefauthors}%
Menictas, M.%
\BCBT {}\ \BBA {} Wand, M.P.%
\end{APACrefauthors}%
\unskip\
\newblock
\APACrefYearMonthDay{2015}{}{}.
\newblock
{\BBOQ}\APACrefatitle {Variational Inference for Heteroscedastic Semiparametric
  Regression} {Variational inference for heteroscedastic semiparametric
  regression}.{\BBCQ}
\newblock
\APACjournalVolNumPages{Australian \& New Zealand Journal of
  Statistics}{57}{1}{119-138}.
\newblock

\newblock

\PrintBackRefs{\CurrentBib}

\bibitem [\protect \citeauthoryear {%
Ong%
\ \protect \BOthers {.}}{%
Ong%
\ \protect \BOthers {.}}{%
{\protect \APACyear {2017}}%
}]{%
Ong2017}
\APACinsertmetastar {%
Ong2017}%
\begin{APACrefauthors}%
Ong, V.M.H.%
, Mensah, D.K.%
, Nott, D.J.%
, Jo, S.%
, Park, B.%
\BCBL {} Choi, T.%
\end{APACrefauthors}%
\unskip\
\newblock
\APACrefYearMonthDay{2017}{}{}.
\newblock
{\BBOQ}\APACrefatitle {{A variational Bayes approach to a semiparametric
  regression using Gaussian process priors}} {{A variational Bayes approach to
  a semiparametric regression using Gaussian process priors}}.{\BBCQ}
\newblock
\APACjournalVolNumPages{Electronic Journal of Statistics}{11}{2}{4258 -- 4296}.
\newblock

\newblock

\PrintBackRefs{\CurrentBib}

\bibitem [\protect \citeauthoryear {%
Paisley%
, Blei%
\BCBL {}\ \BBA {} Jordan%
}{%
Paisley%
\ \protect \BOthers {.}}{%
{\protect \APACyear {2012}}%
}]{%
paisley2012}
\APACinsertmetastar {%
paisley2012}%
\begin{APACrefauthors}%
Paisley, J.W.%
, Blei, D.M.%
\BCBL {} Jordan, M.I.%
\end{APACrefauthors}%
\unskip\
\newblock
\APACrefYearMonthDay{2012}{}{}.
\newblock
{\BBOQ}\APACrefatitle {Variational {B}ayesian Inference with Stochastic Search}
  {Variational {B}ayesian inference with stochastic search}.{\BBCQ}
\newblock
 \APACrefbtitle {Proceedings of the 29th International Conference on Machine
  Learning (ICML-12)} {Proceedings of the 29th international conference on
  machine learning (icml-12)}\ (\BPGS\ 1367--1374).
\PrintBackRefs{\CurrentBib}

\bibitem [\protect \citeauthoryear {%
Park%
\ \BBA {} Casella%
}{%
Park%
\ \BBA {} Casella%
}{%
{\protect \APACyear {2008}}%
}]{%
park2008}
\APACinsertmetastar {%
park2008}%
\begin{APACrefauthors}%
Park, T.%
\BCBT {}\ \BBA {} Casella, G.%
\end{APACrefauthors}%
\unskip\
\newblock
\APACrefYearMonthDay{2008}{}{}.
\newblock
{\BBOQ}\APACrefatitle {The {B}ayesian lasso} {The {B}ayesian lasso}.{\BBCQ}
\newblock
\APACjournalVolNumPages{Journal of the American Statistical
  Association}{103}{482}{681--686}.
\newblock

\newblock

\PrintBackRefs{\CurrentBib}

\bibitem [\protect \citeauthoryear {%
Polson%
, Scott%
\BCBL {}\ \BBA {} Windle%
}{%
Polson%
\ \protect \BOthers {.}}{%
{\protect \APACyear {2014}}%
}]{%
polson2014}
\APACinsertmetastar {%
polson2014}%
\begin{APACrefauthors}%
Polson, N.G.%
, Scott, J.G.%
\BCBL {} Windle, J.%
\end{APACrefauthors}%
\unskip\
\newblock
\APACrefYearMonthDay{2014}{}{}.
\newblock
{\BBOQ}\APACrefatitle {The {B}ayesian bridge} {The {B}ayesian bridge}.{\BBCQ}
\newblock
\APACjournalVolNumPages{Journal of the Royal Statistical Society: Series B
  (Statistical Methodology)}{76}{4}{713--733}.
\newblock

\newblock

\PrintBackRefs{\CurrentBib}

\bibitem [\protect \citeauthoryear {%
Ranganath%
, Gerrish%
\BCBL {}\ \BBA {} Blei%
}{%
Ranganath%
\ \protect \BOthers {.}}{%
{\protect \APACyear {2014}}%
}]{%
ranganath2014}
\APACinsertmetastar {%
ranganath2014}%
\begin{APACrefauthors}%
Ranganath, R.%
, Gerrish, S.%
\BCBL {} Blei, D.M.%
\end{APACrefauthors}%
\unskip\
\newblock
\APACrefYearMonthDay{2014}{}{}.
\newblock
{\BBOQ}\APACrefatitle {Black Box Variational Inference.} {Black box variational
  inference.}{\BBCQ}
\newblock
 \APACrefbtitle {Proceedings of the Seventeenth International Conference on
  Artificial Intelligence and Statistics.} {Proceedings of the seventeenth
  international conference on artificial intelligence and statistics.}
\PrintBackRefs{\CurrentBib}

\bibitem [\protect \citeauthoryear {%
Rue%
, Martino%
\BCBL {}\ \BBA {} Chopin%
}{%
Rue%
\ \protect \BOthers {.}}{%
{\protect \APACyear {2009}}%
}]{%
rue2009}
\APACinsertmetastar {%
rue2009}%
\begin{APACrefauthors}%
Rue, H.%
, Martino, S.%
\BCBL {} Chopin, N.%
\end{APACrefauthors}%
\unskip\
\newblock
\APACrefYearMonthDay{2009}{}{}.
\newblock
{\BBOQ}\APACrefatitle {Approximate Bayesian inference for latent Gaussian
  models by using integrated nested Laplace approximations} {Approximate
  bayesian inference for latent gaussian models by using integrated nested
  laplace approximations}.{\BBCQ}
\newblock
\APACjournalVolNumPages{Journal of the royal statistical society: Series b
  (statistical methodology)}{71}{2}{319--392}.
\newblock

\newblock

\PrintBackRefs{\CurrentBib}

\bibitem [\protect \citeauthoryear {%
Silverman%
\ \BBA {} Green%
}{%
Silverman%
\ \BBA {} Green%
}{%
{\protect \APACyear {1994}}%
}]{%
silv:gree:1994}
\APACinsertmetastar {%
silv:gree:1994}%
\begin{APACrefauthors}%
Silverman, B.W.%
\BCBT {}\ \BBA {} Green, P.J.%
\end{APACrefauthors}%
\unskip\
\newblock
\APACrefYear{1994}.
\newblock
\APACrefbtitle {Nonparametric Regression and Generalized Linear Models}
  {Nonparametric regression and generalized linear models}.
\newblock
\APACaddressPublisher{}{Chapman and Hall (London)}.
\PrintBackRefs{\CurrentBib}

\bibitem [\protect \citeauthoryear {%
Subbotin%
}{%
Subbotin%
}{%
{\protect \APACyear {1923}}%
}]{%
subbotin1923}
\APACinsertmetastar {%
subbotin1923}%
\begin{APACrefauthors}%
Subbotin, M.T.%
\end{APACrefauthors}%
\unskip\
\newblock
\APACrefYearMonthDay{1923}{}{}.
\newblock
{\BBOQ}\APACrefatitle {On the law of frequency of error} {On the law of
  frequency of error}.{\BBCQ}
\newblock
\APACjournalVolNumPages{Sbornik: Mathematics}{31}{2}{296--301}.
\newblock

\newblock

\PrintBackRefs{\CurrentBib}

\bibitem [\protect \citeauthoryear {%
Tibshirani%
}{%
Tibshirani%
}{%
{\protect \APACyear {1996}}%
}]{%
tibshirani1996}
\APACinsertmetastar {%
tibshirani1996}%
\begin{APACrefauthors}%
Tibshirani, R.%
\end{APACrefauthors}%
\unskip\
\newblock
\APACrefYearMonthDay{1996}{}{}.
\newblock
{\BBOQ}\APACrefatitle {Regression shrinkage and selection via the lasso}
  {Regression shrinkage and selection via the lasso}.{\BBCQ}
\newblock
\APACjournalVolNumPages{Journal of the Royal Statistical Society: Series B
  (Methodological)}{58}{1}{267--288}.
\newblock

\newblock

\PrintBackRefs{\CurrentBib}

\bibitem [\protect \citeauthoryear {%
Tibshirani%
}{%
Tibshirani%
}{%
{\protect \APACyear {1997}}%
}]{%
tibshirani1997}
\APACinsertmetastar {%
tibshirani1997}%
\begin{APACrefauthors}%
Tibshirani, R.%
\end{APACrefauthors}%
\unskip\
\newblock
\APACrefYearMonthDay{1997}{}{}.
\newblock
{\BBOQ}\APACrefatitle {The lasso method for variable selection in the Cox
  model} {The lasso method for variable selection in the cox model}.{\BBCQ}
\newblock
\APACjournalVolNumPages{Statistics in medicine}{16}{4}{385--395}.
\newblock

\newblock

\PrintBackRefs{\CurrentBib}

\bibitem [\protect \citeauthoryear {%
Wand%
}{%
Wand%
}{%
{\protect \APACyear {2017}}%
}]{%
Wand2017}
\APACinsertmetastar {%
Wand2017}%
\begin{APACrefauthors}%
Wand, M.P.%
\end{APACrefauthors}%
\unskip\
\newblock
\APACrefYearMonthDay{2017}{}{}.
\newblock
{\BBOQ}\APACrefatitle {Fast Approximate Inference for Arbitrarily Large
  Semiparametric Regression Models via Message Passing} {Fast approximate
  inference for arbitrarily large semiparametric regression models via message
  passing}.{\BBCQ}
\newblock
\APACjournalVolNumPages{Journal of the American Statistical
  Association}{112}{517}{137-168}.
\newblock

\newblock

\PrintBackRefs{\CurrentBib}

\bibitem [\protect \citeauthoryear {%
West%
\ \BBA {} Harrison%
}{%
West%
\ \BBA {} Harrison%
}{%
{\protect \APACyear {1997}}%
}]{%
west2006}
\APACinsertmetastar {%
west2006}%
\begin{APACrefauthors}%
West, M.%
\BCBT {}\ \BBA {} Harrison, J.%
\end{APACrefauthors}%
\unskip\
\newblock
\APACrefYear{1997}.
\newblock
\APACrefbtitle {{B}ayesian forecasting and dynamic models} {{B}ayesian
  forecasting and dynamic models}.
\newblock
\APACaddressPublisher{}{Springer Series in Statistics, New
  York:Springer-Verlag}.
\PrintBackRefs{\CurrentBib}

\bibitem [\protect \citeauthoryear {%
S.~Wood%
}{%
S.~Wood%
}{%
{\protect \APACyear {2012}}%
}]{%
wood2012}
\APACinsertmetastar {%
wood2012}%
\begin{APACrefauthors}%
Wood, S.%
\end{APACrefauthors}%
\unskip\
\newblock
\APACrefYearMonthDay{2012}{}{}.
\newblock
{\BBOQ}\APACrefatitle {mgcv: Mixed GAM Computation Vehicle with GCV/AIC/REML
  smoothness estimation} {mgcv: Mixed gam computation vehicle with gcv/aic/reml
  smoothness estimation}.{\BBCQ}
\newblock
\APACjournalVolNumPages{R package version 1.8-1}{}{}{}.
\newblock

\newblock

\PrintBackRefs{\CurrentBib}

\bibitem [\protect \citeauthoryear {%
S.N.~Wood%
}{%
S.N.~Wood%
}{%
{\protect \APACyear {2017}}%
}]{%
wood2017}
\APACinsertmetastar {%
wood2017}%
\begin{APACrefauthors}%
Wood, S.N.%
\end{APACrefauthors}%
\unskip\
\newblock
\APACrefYear{2017}.
\newblock
\APACrefbtitle {Generalized additive models: an introduction with {R}}
  {Generalized additive models: an introduction with {R}}\ (\PrintOrdinal{2nd}\
  \BEd).
\newblock
\APACaddressPublisher{}{Boca Raton, FL: CRC Press}.
\PrintBackRefs{\CurrentBib}

\bibitem [\protect \citeauthoryear {%
Yin%
\ \BBA {} Zhou%
}{%
Yin%
\ \BBA {} Zhou%
}{%
{\protect \APACyear {2018}}%
}]{%
yin2018}
\APACinsertmetastar {%
yin2018}%
\begin{APACrefauthors}%
Yin, M.%
\BCBT {}\ \BBA {} Zhou, M.%
\end{APACrefauthors}%
\unskip\
\newblock
\APACrefYearMonthDay{2018}{}{}.
\newblock
{\BBOQ}\APACrefatitle {Semi-implicit variational inference} {Semi-implicit
  variational inference}.{\BBCQ}
\newblock
 \APACrefbtitle {International Conference on Machine Learning} {International
  conference on machine learning}\ (\BPGS\ 5660--5669).
\PrintBackRefs{\CurrentBib}

\bibitem [\protect \citeauthoryear {%
Zeiler%
}{%
Zeiler%
}{%
{\protect \APACyear {2012}}%
}]{%
zeiler2012}
\APACinsertmetastar {%
zeiler2012}%
\begin{APACrefauthors}%
Zeiler, M.D.%
\end{APACrefauthors}%
\unskip\
\newblock
\APACrefYearMonthDay{2012}{}{}.
\newblock
{\BBOQ}\APACrefatitle {Adadelta: an adaptive learning rate method} {Adadelta:
  an adaptive learning rate method}.{\BBCQ}
\newblock
\APACjournalVolNumPages{arXiv preprint arXiv:1212.5701}{}{}{}.
\newblock

\newblock

\PrintBackRefs{\CurrentBib}

\bibitem [\protect \citeauthoryear {%
Zou%
\ \BBA {} Hastie%
}{%
Zou%
\ \BBA {} Hastie%
}{%
{\protect \APACyear {2005}}%
}]{%
zou2005}
\APACinsertmetastar {%
zou2005}%
\begin{APACrefauthors}%
Zou, H.%
\BCBT {}\ \BBA {} Hastie, T.%
\end{APACrefauthors}%
\unskip\
\newblock
\APACrefYearMonthDay{2005}{}{}.
\newblock
{\BBOQ}\APACrefatitle {Regularization and variable selection via the elastic
  net} {Regularization and variable selection via the elastic net}.{\BBCQ}
\newblock
\APACjournalVolNumPages{Journal of the royal statistical society: series B
  (statistical methodology)}{67}{2}{301--320}.
\newblock

\newblock

\PrintBackRefs{\CurrentBib}

\end{thebibliography}


\end{document}